\documentclass[lettersize,journal]{IEEEtran}
\usepackage{amsmath,amsfonts}
\usepackage{algorithmic}
\usepackage{booktabs}
\usepackage{color, colortbl}
\usepackage{dsfont}
\usepackage{microtype}
\usepackage{threeparttable}
\usepackage{soul}
\usepackage{multirow}
\usepackage[linesnumbered,ruled,vlined]{algorithm2e}
\usepackage[none]{hyphenat}
\usepackage{array}
\usepackage[caption=false,font=normalsize,labelfont=sf,textfont=sf]{subfig}
\usepackage{textcomp}
\usepackage{stfloats}
\usepackage{url}
\usepackage{verbatim}
\usepackage{graphicx}
\usepackage{cite}
\usepackage{xspace}
\def\eg{\textit{e.g.}}

\def\etal{\textit{et al. }}

\makeatletter
\newcommand*{\rome}[1]{\expandafter\@slowromancap\romannumeral #1@}
\makeatother

\definecolor{orange}{rgb}{1.0, 0.5, 0.0}

\usepackage[normalem]{ulem}
\definecolor{Gray}{gray}{0.9}
\definecolor{Red}{RGB}{245,220,220}

\begin{document}

\title{Learning Dual Transformers for All-In-One Image Restoration from a Frequency Perspective}

\author{{Jie Chu*, Tong Su* , Pei Liu , Yunpeng Wu, Le Zhang, Zenglin Shi, and Meng Wang, Fellow, IEEE}
\thanks{Jie Chu, Tong Su, Zenglin Shi and Meng Wang are with the Hefei University of Technology. Pei Liu and Yunpeng Wu are with the Zhengzhou University. Le Zhang is with the University of Electronic Science and Technology of China. Jie Chu and Tong Su have equal contributions.}

\thanks{Manuscript received April 19, 2021; revised August 16, 2021.}}

\markboth{Journal of \LaTeX\ Class Files,~Vol.~14, No.~8, August~2021}%
{Shell \MakeLowercase{\textit{et al.}}: A Sample Article Using IEEEtran.cls for IEEE Journals}

\IEEEpubid{0000--0000/00\$00.00~\copyright~2021 IEEE}

\maketitle

\begin{abstract}
This work aims to tackle the all-in-one image restoration task, which seeks to handle multiple types of degradation with a single model. The primary challenge is to extract degradation representations from the input degraded images and use them to guide the model's adaptation to specific degradation types. Building on the insight that various degradations affect image content differently across frequency bands, we propose a new dual-transformer approach comprising two components: a frequency-aware Degradation estimation transformer (Dformer) and a degradation-adaptive Restoration transformer (Rformer). The Dformer captures the essential characteristics of various degradations by decomposing the input into different frequency components. By understanding how degradations affect these frequency components, the Dformer learns robust priors that effectively guide the restoration process. The Rformer then employs a degradation-adaptive self-attention module to selectively focus on the most affected frequency components, guided by the learned degradation representations. Extensive experimental results demonstrate that our approach outperforms existing methods in five representative restoration tasks, including denoising, deraining, dehazing, deblurring, and low-light enhancement. Additionally, our method offers benefits for handling, real-world degradations, spatially variant degradations, and unseen degradation levels.

\end{abstract}

\section{Introduction}
\label{sec:intro}
\IEEEPARstart{I}{mage} restoration aims to reconstruct high-quality images from degraded ones affected by issues like noise, blur, resolution loss, and various corruptions. Over time, this field has found extensive applications in diverse real-world scenarios, spanning general visual perception, medical imaging, and satellite imaging. Prevailing image restoration efforts center on the meticulous design of task-specific approaches and have demonstrated promising results in tasks such as denoising \cite{zhang2017beyond,zhang2018ffdnet,shi2021unsharp,shi2022measuring}, deraining \cite{zhang2018density,chen2021robust,yang2020single,zheng2020single}, and deblurring \cite{hu2021pyramid,kupyn2018deblurgan,rim2022realistic,zhang2022blind}. Despite their success in specific tasks, these approaches often prove inadequate when faced with changes in the degradation task or its severity. This limitation presents significant challenges to their practical use in real-world situations, especially in complex environments. For example, self-driving cars may encounter consecutive or simultaneous challenges, such as rainy and hazy weather. Consequently, it becomes imperative to develop more generalized approaches capable of recovering images from a variety of unknown degradation types and levels.

Recent studies, \eg, \cite{chen2020pre, 9157460}, have tried to handle multiple degradations with a multitask learning framework. This involves processing images with different types of degradation by sharing a common backbone and designing task-specific heads. Despite the success of multitask methods in image restoration, those with shared parameters often face the challenge of task interference and still require degradation prior during testing. To address these drawbacks, all-in-one image restoration has recently received increasing attention \cite{li2022all, zhang2023all, potlapalli2023promptir, ai2024multimodal, conde2024high, 10204770, jiang2023autodir, 10204072}. This task aims to address various degradation tasks within a single model. The key challenges in this framework lie in effectively extracting degradation representations from degraded images and utilizing these representations within the restoration network. In this work, we observe through frequency analysis (Section \ref{sec:frequency}) that different types of degradation affect image content differently across frequency bands. Leveraging this insight, we propose a new dual-transformer approach to tackle these challenges: a Frequency-aware Degradation Estimation Transformer (Dformer) to capture degradation representations and a Degradation-Adaptive Restoration Transformer (Rformer) to utilize these representations for effective image restoration.

Dformer is proposed to estimate degradation representation, as the degradation prior is not available in the all-in-one image restoration task. Traditional degradation estimation methods \cite{zhang2017beyond, ren2016image} often assume a predefined degradation type and estimate degradation level, which makes them less effective in scenarios with multiple unknown degradations. Li et al. \cite{li2022all} suggest obtaining degradation representation using a contrastive learning framework, while Park et al. \cite{park2023all} propose learning a degradation classifier to estimate the type of degradation. Potlapalli et al. \cite{potlapalli2023promptir} utilize prompts to encode degradation-specific information. Unlike these methods, our Dformer captures the essential characteristics of various degradations by decomposing features into different frequency components. By understanding how degradations affect these frequency components, Dformer learns robust priors that effectively guide the restoration process.

Rformer functions as a restoration network. The key challenge in designing such a network lies in developing a dynamic module that adapts to various degradation tasks using guidance from degradation representations. Establishing the correlation between the dynamic module and degradation representation is particularly challenging. Li et al. \cite{li2022all} argue that different degradation tasks necessitate different receptive fields within the restoration network. They designed a dynamic module to adjust the receptive field based on the degradation representation. Park et al. \cite{park2023all} introduced an adaptive discriminative filter-based model to explicitly disentangle the restoration network for multiple degradations. Potlapalli et al. \cite{potlapalli2023promptir} proposed a prompt interaction module to enable dynamic interaction between input features and degradation prompts for guided restoration. In contrast, we recognize that different degradation tasks require the restoration model to focus on distinct frequency components of the degraded image. Rformer adapts to these tasks by employing a frequency-aware self-attention mechanism, which allows it to adaptively focus on the most affected frequency components, leading to enhanced restoration performance.

The contributions of this work are summarized as follows:
\begin{itemize}
    \item We propose Dformer, a new degradation estimation transformer that captures the essence of various degradations by decomposing features into frequency components. This frequency-aware design enables Dformer to learn robust priors, effectively guiding the restoration process.
    \item We propose Rformer, a restoration transformer that adaptively targets the most affected frequency subbands. By prioritizing frequency bands impacted by different degradations, Rformer ensures precise and effective restoration.
    \item We conduct extensive experiments to demonstrate that our approach outperforms the existing methods and analyze its advantages of generalization on five representative restoration tasks, including denoising, deraining, dehazing, deblurring and low-light enhancement.
\end{itemize}
Next, we discuss the related works in Section \ref{sec:related}, introduce our approach in detail in Section \ref{sec:method}, and present our various restoration results in Section \ref{sec:experiments}. 

\section{Related Works}
\label{sec:related}
\subsection{Single degradation image restoration}
Numerous methods have been developed to restore clean images from their degraded counterparts. Particularly, convolutional neural networks (CNNs) have demonstrated remarkable efficacy in tasks like image denoising \cite{Dabov2007ColorID, zhang2017beyond, zhang2018ffdnet, tian2020image}, dehazing \cite{cai2016dehazenet, dong2021fdgan, 5567108,zhou2022fsad}, deraining \cite{Wei_2019_CVPR, ren2019progressive, Kui_2020_CVPR, 8099669}, and deblurring \cite{Cho2021RethinkingCA, 7780549, Cui_Tao_Ren_Knoll_2023,liang2024image}. However, with their potent feature representation capabilities, transformer architectures \cite{NIPS2017_3f5ee243} have emerged as a superior alternative to CNNs in image restoration tasks \cite{Tsai2022Stripformer, liang2021swinir, Zamir2021Restormer, Wang_2022_CVPR, si2022inception, NEURIPS2022_a37fea8e, zhang2023accurate}. For instance, by integrating the hierarchical structure of U-Net \cite{RonnebergerFB15} with the multi-head self-attention mechanism of Transformers, Swin Transformer \cite{liu2021swin} effectively balances efficiency and global modeling capabilities through windows-based self-attention. Consequently, Swin Transformer has become a popular choice as the backbone for low-level image restoration models \cite{liang2021swinir, Zamir2021Restormer, Wang_2022_CVPR, si2022inception}. Liang et al. \cite{liang2021swinir} directly employ Swin Transformer for image restoration tasks, augmenting models with various HQ Image Reconstruction layers to cater to different resolution requirements. Similarly, Wang et al. \cite{Wang_2022_CVPR} enhance the decoder layers by replacing the Feed-Forward Network (FFN) with a Locally-enhanced FFN and introducing a learnable modulator before the input of Multi-head Self-attention, allowing for flexible adjustment of feature maps.

However, these methods tailored to specific tasks often struggle to generalize beyond particular types and severities of image degradation. To overcome this challenge, various network design-based approaches have been introduced for general image restoration, demonstrating effectiveness across diverse degradation types \cite{NEURIPS2020_c6e81542, Zamir2021Restormer, liu2018non, Tai-MemNet-2017, Zamir2021MPRNet, 8099783}. Nevertheless, these approaches still rely on training separate models for different datasets and tasks, adhering to a one-by-one paradigm. In contrast, this paper aims to develop an all-in-one image restoration method capable of handling multiple degradations with a single model.

\subsection{Multiple degradations image restoration}

\textbf{Multi-task Methods}. Recent research \cite{chen2020pre, 9157460} has focused on training a single model to address multiple image restoration tasks simultaneously by incorporating separate modules for each task in parallel at the input and output layers. For example, Chen et al. \cite{chen2020pre} developed distinct heads and tails for various tasks, with only the backbone being shared among them. Li et al. \cite{9157460} introduced a task-specific feature extractor to extract common clean features for different adverse weather conditions. However, these multi-task methods still rely on prior knowledge of specific degradations and are unable to handle unknown degradations.

\textbf{All-in-one methods}. 
In contrast to multi-task methods, all-in-one methods \cite{li2022all, zhang2023all, potlapalli2023promptir, ai2024multimodal, conde2024high, 10204770, jiang2023autodir, 10204072} aim to tackle a broad spectrum of image restoration tasks using a single, unified model, thus eliminating the need for prior knowledge of specific degradations or task-specific designs. Wei et al. \cite{Wei_2021_CVPR} and Li et al. \cite{li2022all} pioneered this approach by introducing a novel method that utilizes contrastive learning to extract degradation representations without prior knowledge of the types and levels of degradations, thereby guiding the restoration process. Potlapalli et al. \cite{potlapalli2023promptir} proposed a universal and efficient plugin module that employs adjustable prompts to encode degradation-specific information without prior information on the degradations. Park et al. \cite{10204770} introduced an adaptive discriminant filter-based degradation classifier to explicitly disentangle the network for multiple degradations. Building on the success of vision-language models \cite{radford2021learning}, recent studies \cite{luo2023controlling,yang2024language,liu2023unifying} have integrated language instructions into unified image restoration approaches. However, these methods still require specialized text design for optimal performance.

Unlike the methods discussed earlier, which primarily operate in the spatial domain, this paper presents an all-encompassing image restoration algorithm that carefully considers the variations in frequency across different tasks, aiming to deliver superior results.


\subsection{Frequency-aware image restoration}
Numerous approaches have emerged to address various low-level vision problems, with a notable emphasis on frequency analysis. Alongside spatial representation learning, frameworks that operate in the frequency domain \cite{xu2020learning, 8803391, DeepRFT2021, cui2023selective, jiang2021focal, 8578797,hsu2023wavelet} primarily aim to bridge the frequency gaps between sharp and degraded image pairs. For instance, Yang et al. \cite{8803391} utilize discrete wavelet transform and inverse discrete wavelet transform to replace down-sampling and up-sampling, facilitating the extraction of edge features. Mao et al. \cite{DeepRFT2021} focus on discerning the differences between blurry and sharp images based on low-frequency and high-frequency components, explicitly segregating the processing of these components using Fast Fourier Transform, and ultimately integrating the residual information from both high and low frequencies. Cui et al. \cite{cui2023selective} introduce a selective frequency module that dynamically separates feature maps into distinct frequency components using theoretically validated filters, enabling the extraction of high-frequency and low-frequency information in a non-explicit and learnable manner. A common strategy among existing methods involves decomposing features into various frequency components through transformation tools such as wavelet transform, Fourier transform, pooling techniques, and conventional filters, and then treating each component individually.

Moreover, several recent studies \cite{park2022vision, wang2022antioversmoothing} have investigated the biases of different modules in the frequency domain. For example, it has been observed that the self-attention mechanism in transformer architectures acts as a low-pass filter, while convolution in CNNs behaves like a high-pass filter. This highlights the importance of frequency separation, as it can help mitigate model biases by handling different frequencies separately. In this study, we delve into the varying frequency objectives across different image restoration tasks. Tasks such as denoising and deraining require the suppression of high-frequency noise, while dehazing and deblurring tasks focus on restoring high-frequency details. By adaptively addressing the inherent frequency biases in self-attention modules of transformer-like models, we propose a new frequency-aware all-in-one image restoration method.

\begin{figure*}[htbp]
  \centering
  \subfloat[Noisy]{\includegraphics[width=0.257\textwidth]{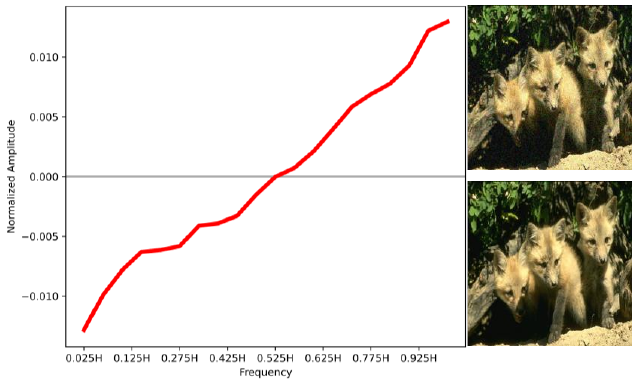}}
  \hfill
  \subfloat[Rainy]{\includegraphics[width=0.247\textwidth]{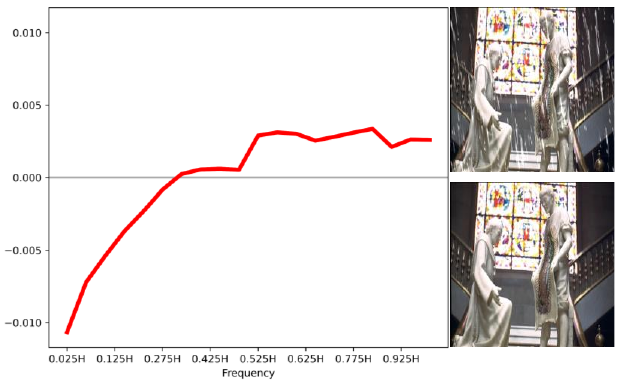}}
  \hfill
  \subfloat[Hazy]{\includegraphics[width=0.247\textwidth]{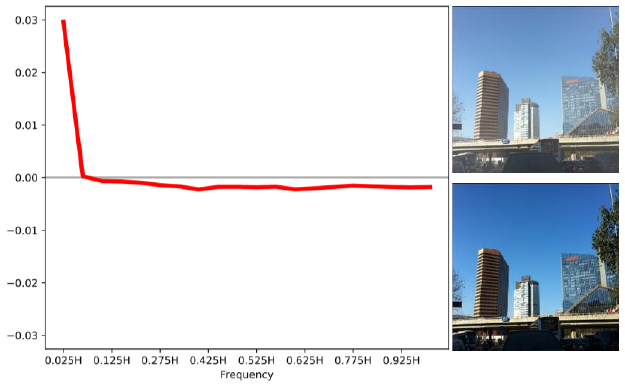}}
  \hfill
  \subfloat[Blurry]{\includegraphics[width=0.247\textwidth]{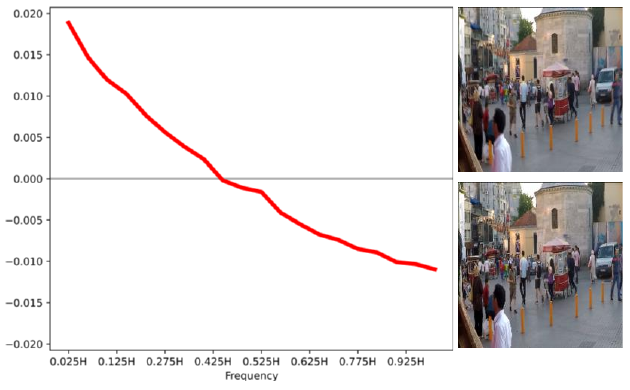}}
  \caption{\textbf{Frequency analysis of various degradation types.} For each degradation type, the images are the discrete normalized spectrogram (left), the degraded image (right top), and the clean image (right bottom). Noisy and rainy images exhibit a reduced proportion of low-frequency components and an increased proportion of high-frequency components compared to their corresponding clean images. In contrast, hazy and blurry images show the opposite trend.}
  \label{fig:frequency}
\end{figure*}
\begin{figure}
    \centering
    \includegraphics[width=0.8\linewidth]{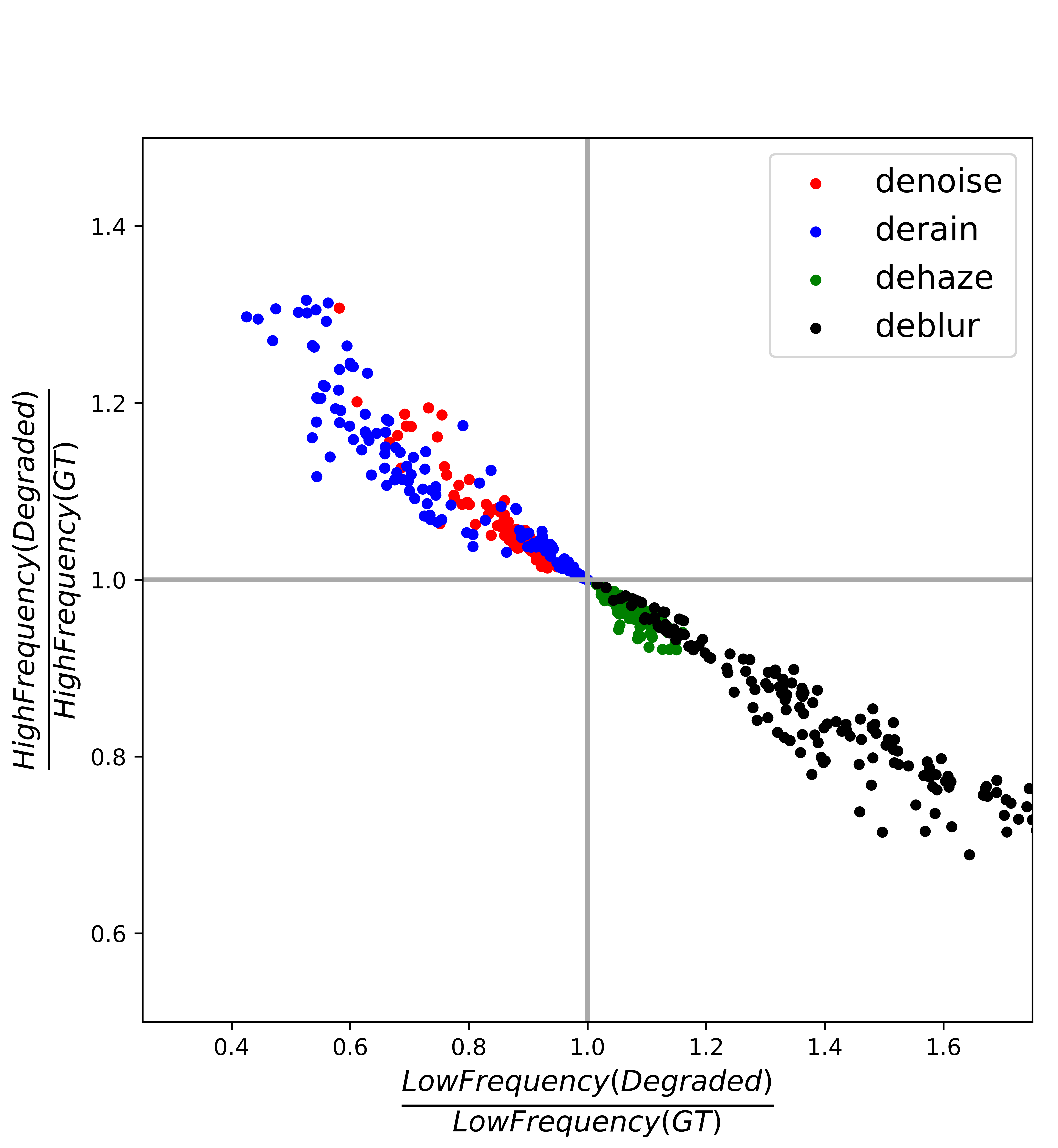}
    \caption{\textbf{Visualizing the ratios of low-frequency to high-frequency components} between the clean and degraded images using 150 samples. The noisy and rainy samples are positioned in the upper-left region, indicating that the degraded images contain more high-frequency components and fewer low-frequency components than their clean counterparts. In contrast, the hazy and blurred samples are found in the lower-right region, reflecting the opposite trend.}
    \label{fig:frequency-1}
\end{figure}

\section{Method}
\label{sec:method}
\begin{figure*}[htbp]
  \centering
  \subfloat[Frequency-aware dynamic transformers]
  {\includegraphics[width=0.62\textwidth]{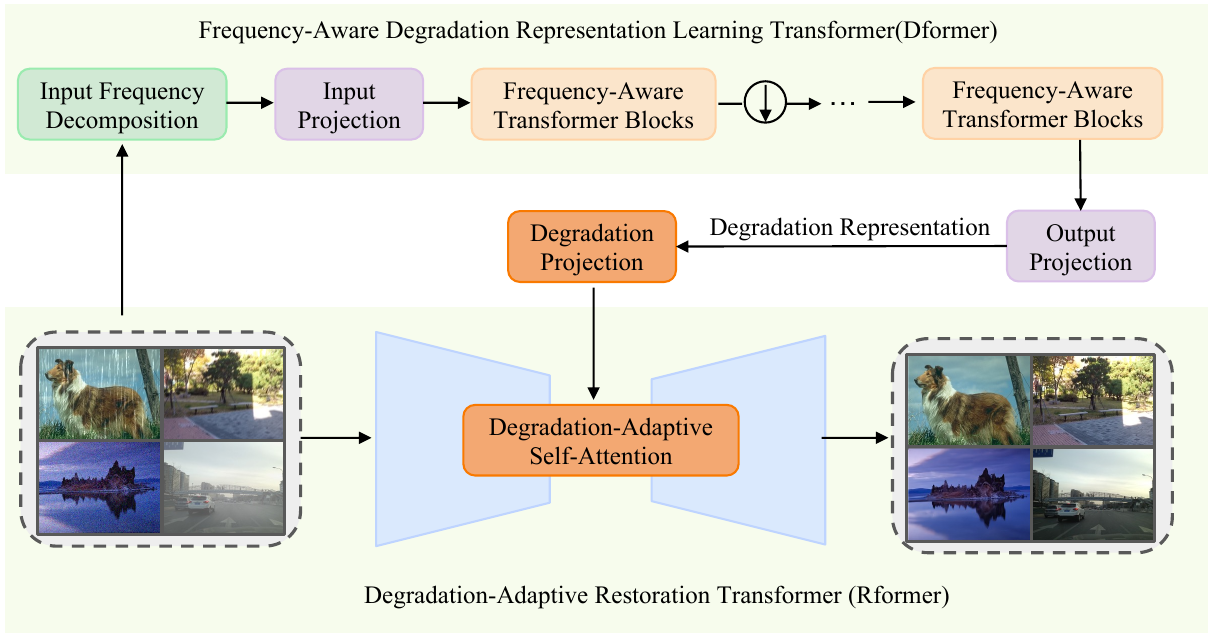}}
  \hfill
  \subfloat[FA-TB]
  {\includegraphics[width=0.21\textwidth]{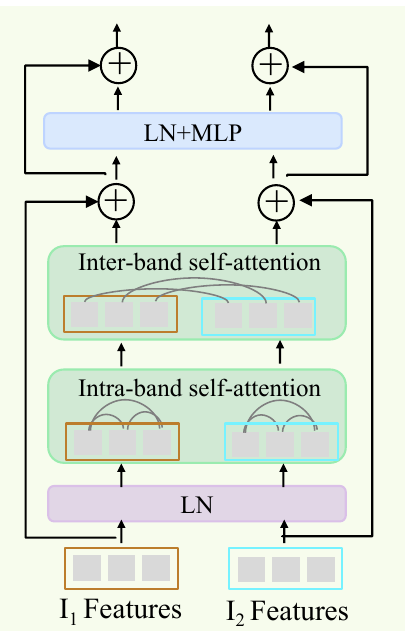}}
  \hfill
  \subfloat[DA-SA]
  {\includegraphics[width=0.118\textwidth]{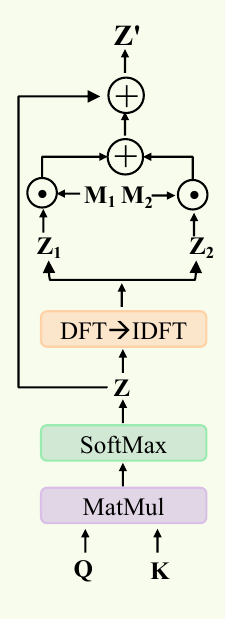}}
  \caption{\textbf{Overview of the proposed methods.} Dformer learns degradation representation and guides Rformer to achieve all-in-one restoration. Input Frequency Decomposition module utilizes DFT and IDFT processes to decompose the input image into multiple frequency-band images. Input Projection module employs a convolution layer to project the input image into the feature maps. Frequency-Aware Transformer Block (FA-TB) is detailed in (b). The image showing a down-arrow within a circle denotes the downsampling layer. Output Projection module includes 2D average pooling and two-layer MLP to refine and project degradation representation. Degradation Projection includes a two-layer MLP. The architecture of Rformer follows Uformer \cite{Wang_2022_CVPR}, but employs a new degradation-adaptive self-attention mechanism as detailed in (c).}
  \label{fig:overview}
\end{figure*}
In this work, we propose a new all-in-one image restoration approach from a frequency perspective. We begin by analyzing how various degradation types exhibit distinct characteristics in the frequency domain. Building on these insights, we then introduce the details of our proposed method.

\subsection{Frequency analysis of various degradation types}
\label{sec:frequency}
In this analysis, we examine four common degradation types that are extensively studied in all-in-one image restoration: noise, rain, haze, and blur. Given the degraded images and their corresponding clean images, We generate spectrograms based on the normalized amplitudes of the residual maps, which are obtained by subtracting the degraded images from the clean images in the frequency domain. To enhance visualization, we divide the spectrograms into 20 equal-width bins, each representing a continuous frequency band. As shown in Fig. \ref{fig:frequency}(a)-(d), we observe that noisy and rainy images exhibit a reduced proportion of low-frequency components and an increased proportion of high-frequency components compared to their corresponding clean images. In contrast, hazy and blurred images show the opposite trend, with a higher concentration of low-frequency components and a lower concentration of high-frequency components. 

We further sample 150 image pairs from each of the four degradation types and categorize the normalized amplitudes into two frequency bands: low-frequency and high-frequency, using a method similar to the one described above. For each image pair, we then compute the ratios of low-frequency to high-frequency components between the clean and degraded images. As shown in Fig. \ref{fig:frequency-1}, we observe that the noisy and rainy image pairs are positioned in the upper-left region, indicating that the degraded images contain more high-frequency components and fewer low-frequency components than their clean counterparts. In contrast, the hazy and blurred image pairs are found in the lower-right region, reflecting the opposite trend.

Building on the analysis above, we conclude that noise and rain primarily degrade images in the high-frequency bands, while haze and blur predominantly affect the low-frequency bands. To address this for all-in-one image restoration, we propose a frequency-aware degradation estimation transformer (Dformer), which captures the key characteristics of various degradations by decomposing inputs into distinct frequency components. By understanding the impact of these degradations on different frequency bands, Dformer learns robust priors that effectively guide the restoration process. We then introduce a degradation-adaptive restoration transformer (Rformer), which incorporates a degradation-adaptive self-attention module. This module enables the model to dynamically focus on the most affected frequency bands, using the degradation representations learned by Dformer. In the following sections, we provide detailed explanations of the architectures and optimizations for Dformer and Rformer. An overview of the proposed approach is shown in Fig. \ref{fig:overview}(a).

\subsection{Frequency-aware degradation estimation transformer}
Dformer constructs a hierarchical encoder network based on the architecture of the Swin Transformer \cite{liu2021swin}, as illustrated in Fig. \ref{fig:overview} (a). Dformer incorporates two key designs: 1) Input Frequency Decomposition Module decomposes the input degraded image into distinct frequency bands. 2) Frequency-aware Swin Transformer Block performs self-attention both within and between these frequency bands, effectively learning degradation representations. 

\textbf{Input frequency decomposition module. } 
Given the RGB degraded image $\boldsymbol{I} \in \mathbb{R}^{3 \times H \times W}$, the module first performs a 2D discrete Fourier transform (DFT) to obtain the Fourier spectrum of $\boldsymbol{I}$. Then the Fourier spectrum of $k$-th frequency band, denoted as $\operatorname{F-Band}_k(\boldsymbol{I}) \in \mathbb{C}^{H \times W}$, can be obtained by:
\begin{align}\label{eq-2}
    \begin{split}
\operatorname{F-Band}_k(\boldsymbol{I})=\begin{cases} {{\mathcal{F}(\boldsymbol{I})_{ij},}} & {{\mathrm{i f ~} |i-\lfloor{\frac n 2}\rfloor|,|j-\lfloor{\frac n 2}\rfloor| \in [l_k,r_k]}} \\ {{0,}} &\mathrm{ otherwise} \\ \end{cases}, 
    \end{split}
\end{align}
where $\mathcal{F} : \mathbb{R}^{n \times n} \to \mathbb{C}^{n \times n}$ denote the 2D DFT. $l_k$ and $r_k$ denote the minimum and maximum frequencies for each band, respectively. The frequency range is divided into $L$ bands, where the first band only contains the direct current (DC) component (i.e., $l_1=r_1=0$), and the remaining bands divide the entire frequency range equally. The Fourier spectrum of each frequency band is transformed back to the spatial domain using the 2D inverse DFT, denoted by ${\mathcal{F}}^{-1} : \mathbb{C}^{n \times n} \to \mathbb{R}^{n \times n}$:
\begin{align}
   \label{eq-5}
    \begin{split}
        \boldsymbol{I_k}={{\mathcal{F}}^{-1}(\operatorname{F-Band}_k(\boldsymbol{I})).}
    \end{split}
\end{align}
where ${\mathcal{F}}^{-1} : \mathbb{C}^{n \times n} \to \mathbb{R}^{n \times n}$ denote the 2D inverse DFT.

\textbf{Frequency-aware Transformer block.} 
The Swin Transformer block employs a shifted window-based self-attention mechanism to efficiently capture both local and global contextual information. Unlike the original Swin Transformer block, which processes a single input image $\boldsymbol{I}$, our enhanced block processes $L$ input images, ${\boldsymbol{I}_1, \boldsymbol{I}_2, \ldots, \boldsymbol{I}_L}$, derived from the input frequency decomposition module. To enable the Swin Transformer block to handle multiple frequency-band inputs and fully leverage their contents, we introduce a new frequency-aware transformer block. This block incorporates new designs in self-attention mechanisms, positional encoding, and masking techniques.

We introduce Intra- and Inter-Band shifted window-based self-attention mechanisms to facilitate adaptive interactions within and between frequency bands, as illustrated in Fig. \ref{fig:overview} (b). \textbf{Intra-band self-attention} facilitates interactions among distinct pixels within each frequency band, essentially performing self-attention computations independently for each band within the Swin Transformer block. This method ensures complete isolation between different frequency bands, focusing exclusively on intra-band interactions. On the other hand, \textbf{inter-band self-attention} explicitly manages interactions across different frequency bands. Utilizing a window-based strategy, it computes self-attention between pixels from different frequency bands within the same spatial window. This approach allows for a more detailed examination of frequency disparities within localized regions.

To adapt the relative positional encoding and window shifting mechanism within the Swin Transformer block to variations in token count and dimensions, we propose integrating a one-dimensional absolute frequency domain positional encoding alongside the original two-dimensional relative spatial positional encoding. Additionally, to facilitate the window shifting mechanism, we introduce an enhanced masking mechanism. This ensures that interactions occur exclusively among tokens within spatially adjacent shifted windows that meet the frequency criteria for both intra- and inter-band self-attention.

\textbf{Degradation estimation}. Given the RGB degraded image $\boldsymbol{I}$, the input frequency decomposition module generates $L$ new images $\{\boldsymbol{I}_1,\boldsymbol{I}_2,\ldots,\boldsymbol{I}_L\}$, each corresponding to a different frequency band of the degraded image $\boldsymbol{I}$. These images $\{\boldsymbol{I}_1,\boldsymbol{I}_2,\ldots,\boldsymbol{I}_L\}$ are then passed through a shared $3 \times 3$ convolutional layer to extract low-level features. The extracted features are subsequently processed through $K$ shared encoder stages. Each stage consists of $N$ frequency-aware Swin Transformer blocks and a downsampling layer, except for the last stage. 
The downsampling layer reshapes the flattened features into 2D spatial feature maps, doubles the number of channels, and downsamples the feature maps using a $4 \times 4$ convolutional layer with a stride of 2. After the $K$ encoder stages, we use an output projection, which includes 2D average pooling and two-layer MLP, to generate a degradation representation vector. 

\subsection{Degradation-adaptive restoration transformer}
After obtaining the degradation representations, we incorporate them into a restoration transformer (Rformer), as illustrated in Fig. \ref{fig:overview} (a). Rformer is designed as a general framework adaptable to various transformer-based image restoration architectures, but employs a new degradation-adaptive self-attention mechanism to adaptively focus on the most affected frequency bands, guided by the acquired degradation representations. In this work, we adopt the Uformer \cite{Wang_2022_CVPR} as a specific instantiation of Rformer.

\textbf{Degradation adaptive self-attention.} Generally, a self-attention map $\boldsymbol{z}$ in Transformer can be computed as:
\begin{equation}
    \boldsymbol{z} = softmax(\frac{\boldsymbol{X} \boldsymbol{W}_Q(\boldsymbol{X} \boldsymbol{W}_K)^T}{\sqrt{d}})
\end{equation}
where $\boldsymbol{X} \in \mathbb{R}^{n \times d}$ denotes the feature maps, $n$ and $d$ denote the number of tokens and the feature dimension. $\boldsymbol{W}_K, \boldsymbol{W}_Q \in \mathbb{R}^{d \times d}$ are the weight matrices for key and query, respectively. Formally, the frequency band of the attention maps $\boldsymbol{z}$ can be obtained by Eq. \ref{eq-2}. $\operatorname{F-Band}_k(\boldsymbol{z})$ denotes the $k$-th frequency band partitioned from the attention maps. After frequency decomposition, the frequency is performed as follows:
\begin{align}\label{eq-3}
    \begin{split}    \boldsymbol{z}^{'}=\boldsymbol{z}+\sum_{k>1}^{L}\boldsymbol{M}_{k-1}{\mathcal{F}}^{-1}(\operatorname{F-Band}_k(\boldsymbol{z})),
    \end{split}
\end{align}
where $\boldsymbol{z}^{'}$ denotes the modulated attention map, $\boldsymbol{M}_k$ denotes the modulation ratio for the $k$-th frequency band. $\boldsymbol{M}_k$ is initialized to zero, ensuring that \(\boldsymbol{z}' = \boldsymbol{z}\), indicating that no modulation is applied initially. The direct component serves as a baseline for modulating other frequency bands since it remains unscaled. Put plainly, the modulation process entails segmenting the attention map into distinct frequency bands and independently modulating each band using specific modulation ratios. The frequency modulation ratios $M=\{\boldsymbol{M}_0, \boldsymbol{M}_1,\ldots,\boldsymbol{M}_L\}$ are embedded using a degradation projection implemented by a two-layer MLP, which takes the degradation representations $\boldsymbol{d}$, derived from Dformer, as input.

\subsection{Composite training loss}
The training of our approach is carried out in two distinct stages. Initially, Dformer is trained to learn degradation representations with a contrastive learning loss $\mathcal{L}_{cl}$. We consider $\boldsymbol{d}$ as the degradation representation of the anchor sample, $\boldsymbol{d}^{+}$ and $\boldsymbol{d}^{-}$ as the degradation representation of positive and negative samples obtained through the MoCo framework, where positive samples and the anchor sample come from the same degraded image, while negative samples comes from other degraded images. The samples are random patches of size $128 \times 128$ from the degraded images and undergo random data augmentation. $\mathcal{L}_{cl}$ is defined by:
\begin{align}\label{eq-6}
    \begin{split}
    \mathcal{L}_{cl}=-\operatorname{log} \frac{\operatorname{exp} ( \boldsymbol{d} \cdot \boldsymbol{d}^{+} / \tau)} {\sum_{\boldsymbol{d}^{-} \in Queue}\operatorname{exp} ( \boldsymbol{d} \cdot \boldsymbol{d}^{-} / \tau)}
    \end{split}
\end{align}
where $Queue$ represents the negative sample queue in the MoCo framework, and $\tau$ denotes the temperature hyperparameter.

In the second stage, we train the Dformer and Rformer together by using a composite loss function. This loss function comprises two distinct components:
\begin{equation}
    \mathcal{L} = \mathcal{L}_{cl} + \mathcal{L}_{rec},
\end{equation}
where $\mathcal{L}_{rec}=|\hat{\boldsymbol{I}}-\boldsymbol{Y}|$ is a L1 loss. Here, $\hat{\boldsymbol{I}}$ denotes the recovered image through Rformer, and $\boldsymbol{Y}$ is the corresponding clean image.

\section{Experiments and Results}
\label{sec:experiments}
In this section, we comprehensively evaluate and analyze our proposed method across five tasks: denoising, deraining, dehazing, deblurring, and low-light enhancement. We start by detailing the experimental setup, followed by presenting both qualitative and quantitative results from benchmark tests, which demonstrate the effectiveness of our method for all-in-one image restoration. Additionally, we conduct ablation studies to validate the contribution of each component of our approach. Finally, we provide further analysis to highlight additional benefits of our method.

\subsection{Experimental Setup}
\textbf{Datasets.}
To assess the effectiveness of the proposed approaches in multi-degradation restoration, we mainly employ seven datasets: BSD400 \cite{martin2001database}, BSD68 \cite{martin2001database}, WED \cite{ma2016waterloo}, and Urban100 \cite{huang2015single} for image denoising, Rain100L \cite{yang2019joint} for image deraining, RESIDE \cite{li2018benchmarking} for image dehazing, GoPro \cite{nah2017deep} for image deblurring, and LOL dataset \cite{wei2018deep} for low-light enhancement. Detailed information on the datasets is provided in Table \ref{tab:dataset}. Following Li \etal \cite{li2022all}, the noisy images for image denoising are generated by manually adding white Gaussian noise to the clean images at three corruption levels, namely, $\sigma=15,25,50$.
\begin{table}[!h]
\small
\centering
\caption{\textbf{The details of datasets used for training and testing.} The number in each bracket indicates the quantity of images or image pairs contained in the corresponding dataset.}
\label{tab:dataset}
\resizebox{0.99\linewidth}{!}{
\begin{tabular}{llllll}
\toprule
Degradation type &Training & Testing\\
\hline
 Denoise&BSD400 \cite{martin2001database} (400), WED  \cite{ma2016waterloo} (4744)&BSD68 \cite{martin2001database} (68)\\
 Derain&Rain100L-train \cite{yang2019joint} (200) &Rain100L-test \cite{yang2019joint} (100) \\
 Dehaze&RESIDE-OTS \cite{li2018benchmarking} (72135) &RESIDE-SOTS \cite{li2018benchmarking} (500) \\
Deblur&GoPro-train \cite{nah2017deep} (2103) &GoPro-test \cite{nah2017deep} (1111)\\
Low-light&LOL-train \cite{wei2018deep} (485) &LOL-test \cite{wei2018deep} (15)\\
\bottomrule
\end{tabular}
}
\end{table}

\textbf{Implementation details.}
The Rformer in this paper represents a general image restoration architecture based on the Transformer framework, adaptable to various specific Transformer implementations. In experiments, We leverage Uformer \cite{Wang_2022_CVPR} as the Rformer, due to its extensively validated performance and computational efficiency across various restoration tasks. Our approach is implemented using the PyTorch framework and trained on a single NVIDIA A800 GPU. 

During training, we utilize the AdamW optimizer to optimize the network parameters, following the configuration settings outlined in AirNet \cite{li2022all}. The training phase takes 1000 epochs: the initial 100 epochs are dedicated to training the Encoder using Contrastive Loss optimization for warm-up, while the subsequent 900 epochs focus on optimizing the entire network (both Encoder and backbone) using total loss optimization. In the first 100 epochs, the learning rate is initialized at 3e-4 and is then reduced to 3e-5 after the 60th epoch. For the subsequent 900 epochs, the learning rate is initialized at 1e-4 and is halved every 125 epochs to facilitate convergence. During training, we fix the image patch size at $128 \times 128$ and apply random data augmentations to the input image pairs, in line with the standard practices of contrastive learning methodologies. The batch size is set to $400 \times N$, where $N$ represents the number of degradation types. We set the number of frequency bands to $L=2$ across all experiments to achieve an optimal balance between efficiency and performance, as demonstrated by the experiments in Section \ref{sec:analysis-frequency bands}.

During testing, to ensure consistency in patch size for our Transformer-like architecture, we segment images into multiple $128 \times 128$ patches, following the same procedure as in training. For images larger than $640 \times 360$, we divide them into non-overlapping patches. For images smaller than or equal to $640 \times 360$, we adopt the strategy from SwinIR, sampling a patch every $64 \times 64$ pixels. This approach effectively balances border artifacts and computational efficiency.

\textbf{Metrics.}
In line with Li \etal \cite{li2022all}, we employ two widely used metrics for quantitative comparisons: Peak Signal-to-Noise Ratio (PSNR) and Structural Similarity (SSIM). A superior performance is indicated by higher values of these metrics.

\subsection{Comparison to the state-of-the-art}
We first perform a comparison to the state-of-the-art in the conventional “noise-rain-haze" setting to showcase the superiority of our approach. Our comparison encompasses four single-degradation image restoration techniques, namely BDRNet \cite{TIAN2020461}, LP-Net \cite{fu2019lightweight}, FDGAN \cite{dong2021fdgan}, and MPRNet \cite{Zamir2021MPRNet}, alongside the multi-task method for multiple degradation image restoration, DL \cite{8750830}. We are also specifically evaluating two specialized all-in-one methods, AirNet \cite{li2022all} and PromptIR \cite{potlapalli2023promptir}. 

The results, as shown in Table \ref{tab:sota}, highlight the superiority of all all-in-one methods over other baselines for single degradation, underscoring their ability to address various unknown degradations within a unified framework. Notably, our approach demonstrates even better performance compared to other all-in-one methods. Specifically, we surpass AirNet \cite{li2022all} across all tasks, achieving an average performance improvement of $1.12$ dB PSNR and $0.011$ SSIM. Furthermore, we outperform PromptIR \cite{potlapalli2023promptir} in denoising and deraining tasks, with an average performance improvement of $0.26$ dB PSNR and $0.008$ SSIM. 

\begin{figure*}[htbp]
  \centering
  \subfloat[Degraded]{\includegraphics[width=0.195\textwidth]{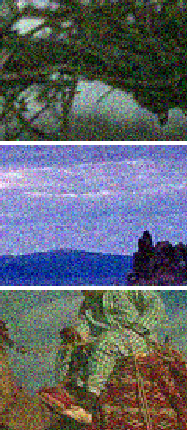}}
  \hfill
  \subfloat[AirNet]{\includegraphics[width=0.195\textwidth]{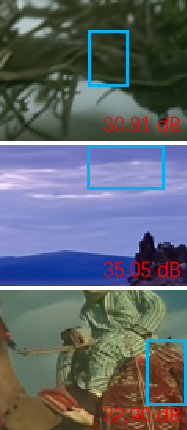}}
  \hfill
  \subfloat[PromptIR]{\includegraphics[width=0.195\textwidth]{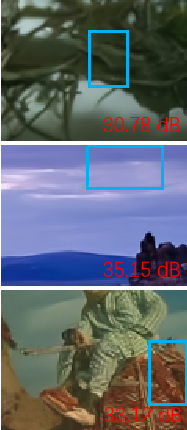}}
  \hfill
  \subfloat[Ours]{\includegraphics[width=0.195\textwidth]{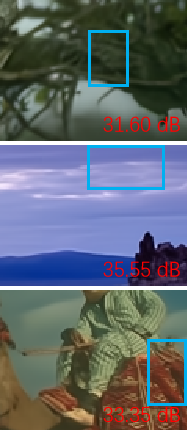}}
  \hfill
  \subfloat[Clear]{\includegraphics[width=0.195\textwidth]{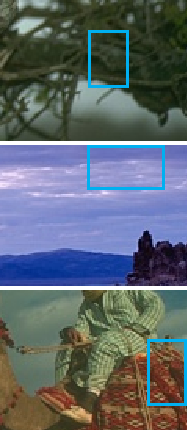}}
  \caption{\textbf{The performance of various methods on denoising ($\sigma=25$) tasks.} From the regions masked by the blue rectangles, we observe our method better preserves edge details (best viewed digitally).}
  \label{fig:denoising}
\end{figure*}

\begin{figure*}[htbp]
  \centering
  \subfloat[Degraded]{\includegraphics[width=0.195\textwidth]{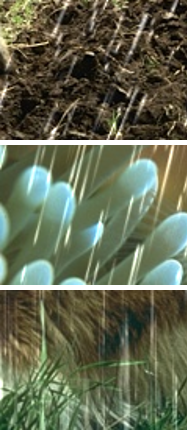}}
  \hfill
  \subfloat[AirNet]{\includegraphics[width=0.195\textwidth]{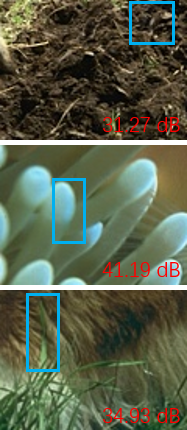}}
  \hfill
  \subfloat[PromptIR]{\includegraphics[width=0.195\textwidth]{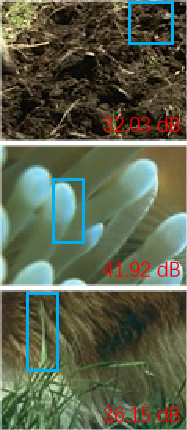}}
  \hfill
  \subfloat[Ours]{\includegraphics[width=0.195\textwidth]{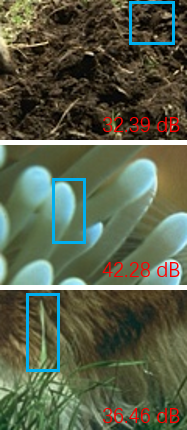}}
  \hfill
  \subfloat[Clear]{\includegraphics[width=0.195\textwidth]{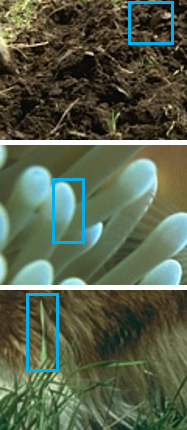}}
  \caption{\textbf{The performance of various methods on deraining tasks.} From the regions masked by the blue rectangles, we observe our method performs well, especially when recovering high-frequency details (best viewed digitally).}
  \label{fig:deraining}
\end{figure*}

\begin{figure*}[htbp]
  \centering
  \subfloat[Degraded]{\includegraphics[width=0.195\textwidth]{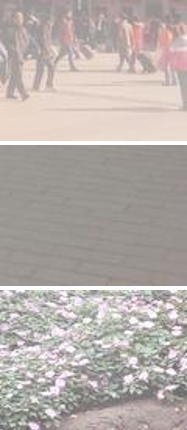}}
  \hfill
  \subfloat[AirNet]{\includegraphics[width=0.195\textwidth]{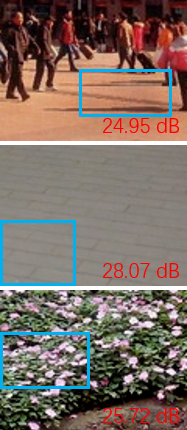}}
  \hfill
  \subfloat[PromptIR]{\includegraphics[width=0.195\textwidth]{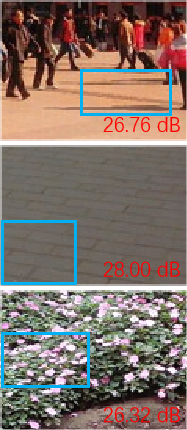}}
  \hfill
  \subfloat[Ours]{\includegraphics[width=0.195\textwidth]{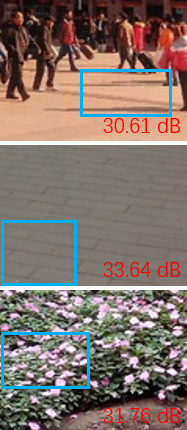}}
  \hfill
  \subfloat[Clear]{\includegraphics[width=0.195\textwidth]{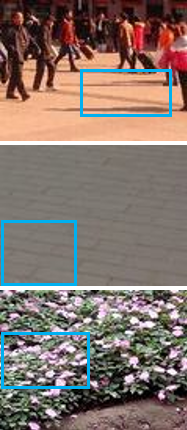}}
  \caption{\textbf{The performance of various methods on dehazing tasks.} From the regions masked by the green rectangles, we observe our method achieves better color fidelity (best viewed digitally).}
  \label{fig:dehazing}
\end{figure*}

Qualitative examples are presented in Fig. \ref{fig:denoising},\ref{fig:deraining}, and \ref{fig:dehazing}. Compared to AirNet \cite{li2022all} and PromptIR \cite{potlapalli2023promptir}, our approach better preserves edge details when performing denoising and deraining, and achieves better color fidelity when performing dehazing.

\begin{table*}[t!]
\centering
\caption{\textbf{Comparison to the state-of-the-art} on the conventional “noise-rain-haze” setting. Existing all-in-one methods surpass other baselines designed for single degradation tasks, whereas our approach showcases superior performance. }
\resizebox{0.99 \linewidth}{!}{
\begin{tabular}{@{}lccccccccccccr@{}}
\toprule
 \multirow{2}{*}{ Method } & \multicolumn{3}{c}{ Denoise } & Derain & Dehaze & \multirow{2}{*}{ Average } \\
\cmidrule(lr){2-4} \cmidrule(lr){5-5} \cmidrule(lr){6-6} 
& BSD68 $(\sigma=15)$ & BSD68 $(\sigma=25)$ & BSD68 $(\sigma=50)$ & Rain100L & SOTS\\
\hline BRDNet\cite{TIAN2020461}&32.26/0.898&29.76/0.836&26.34/0.693&27.42/0.895&23.23/0.895&27.80/0.843\\
LPNet\cite{8953950}&26.47/0.778&24.77/0.748&21.26/0.552&24.88/0.784&20.84/0.828&23.64/0.738\\
FDGAN\cite{dong2021fdgan}&30.25/0.910&28.81/0.868&26.43/0.776&29.89/0.933&24.71/0.929&28.02/0.883\\
MPRNet\cite{Zamir2021MPRNet}&33.54/0.927&30.89/0.880&27.56/0.779&33.57/0.954&25.28/0.955&30.17/0.899\\
DL\cite{8750830}&33.05/0.914&30.41/0.861&26.90/0.740&32.62/0.931&26.92/0.931&29.98/0.876\\
AirNet\cite{li2022all}&33.92/0.933&31.26/0.888&28.00/0.797&34.90/0.968&27.94/0.962&31.20/0.910\\
PromptIR\cite{potlapalli2023promptir}&33.98/0.933&31.31/0.888&28.06/0.799&36.37/0.972&\textbf{30.58}/\textbf{0.974}&32.06/0.913\\
\hline 
\rowcolor{Gray}
\textit{Ours} & \textbf{34.59}/\textbf{0.941} & \textbf{31.83}/\textbf{0.900} & \textbf{28.46}/\textbf{0.814} & \textbf{37.50}/\textbf{0.980} & 29.20/0.972 & \textbf{32.32}/\textbf{0.921} \\
\bottomrule
\end{tabular}}
\label{tab:sota}
\end{table*}

\subsection{Comparison on the number of degradation types}
In this experiment, we conduct a comparative analysis between the proposed method, AirNet \cite{li2022all}, and PromptIR \cite{potlapalli2023promptir} across different numbers of degradations to assess the stability of our approach.
The experimental results in Table \ref{tab:multiple} shows that as the number of degradation types increases, the network's ability to restore images diminishes, resulting in a performance decline. Notably, both AirNet and PromptIR experience significant performance degradation when tasked with handling multiple degradations simultaneously. 
For instance, the PSNR for deraining drops from 38.31 dB to 34.7 dB for AirNet, and from 39.32 dB to 36.14 dB for PromptIR, as the number of combined degradation types increases from 2 to 4. This decline in performance occurs due to potential conflicts between different tasks during joint learning, which AirNet and PromptIR struggle to manage effectively. 

\begin{figure*}[htbp]
  \centering
  \subfloat[Degraded]{\includegraphics[width=0.195\textwidth]{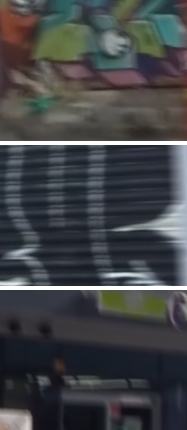}}
  \hfill
  \subfloat[AirNet]{\includegraphics[width=0.195\textwidth]{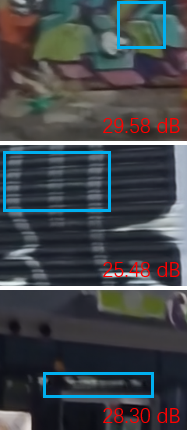}}
  \hfill
  \subfloat[PromptIR]{\includegraphics[width=0.195\textwidth]{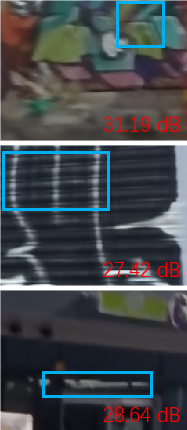}}
  \hfill
  \subfloat[Ours]{\includegraphics[width=0.195\textwidth]{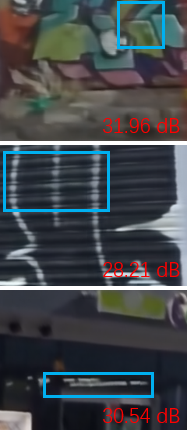}}
  \hfill
  \subfloat[Clear]{\includegraphics[width=0.195\textwidth]{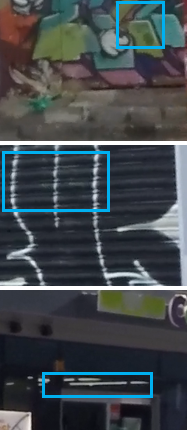}}
  \caption{\textbf{The performance of various methods on deblurring tasks.} From the regions masked by the blue rectangles, we observe our method better recovers high-frequency edges (best viewed digitally).}
  \label{fig:deblurring}
\end{figure*}

In contrast, our method explicitly addresses task disparities in frequency domain through frequency-aware dynamic transformers. Consequently, our method experiences a milder drop of only 1.43 dB, from 39.51 dB PSNR to 37.35 dB PSNR, showcasing superior stability across varying numbers of degradation types. 
When four types of combined degradations are present, qualitative examples from deblurring tasks are shown in Fig. \ref{fig:deblurring}. Our approach demonstrates superior edge detail preservation compared to AirNet \cite{li2022all} and PromptIR \cite{potlapalli2023promptir}.

\begin{table*}[t!]
\centering
\caption{\textbf{Comparison on the number of degradation types.} As the number of combined degradation types increases, our proposed approach demonstrates superior performance stability compared to AirNet and PromptIR.}
\resizebox{0.99 \linewidth}{!}{
\begin{tabular}{@{}lccccccccccccr@{}}
\toprule
\multirow{2}{*}{ D-Types } & \multirow{2}{*}{ Method } & \multicolumn{3}{c}{ Denoise } & Derain & Dehaze & Deblur \\
\cmidrule(lr){3-5} \cmidrule(lr){6-6} \cmidrule(lr){7-7}  \cmidrule(lr){8-8} 
&& BSD68 $(\sigma=15)$ & BSD68 $(\sigma=25)$ & BSD68 $(\sigma=50)$ & Rain100L & SOTS & GOPro \\
\hline \multirow{2}{*}{ 1 } 
& AirNet & 34.14/0.936 & 31.49/0.893 & 28.23/0.806 & - & - & - \\
& PromptIR & 34.34/0.940 & 31.71/0.900 & 28.49/0.813 & - & - & - \\
& \textit{Ours} & \textbf{34.74}/\textbf{0.943} & \textbf{31.98}/\textbf{0.903} & \textbf{28.66}/\textbf{0.820} & - & - & - \\
\hline \multirow{2}{*}{ 2} 
& AirNet & 34.11/0.935 & 31.46/0.892 & 28.19/0.804 & 38.31/0.982 & - & - \\
& PromptIR & 34.26/0.937 & 31.61/0.895 & 28.37/0.810 & 39.32/0.986 & - & - \\
& \textit{Ours} & \textbf{34.69}/\textbf{0.942} & \textbf{31.93}/\textbf{0.902} & \textbf{28.59}/\textbf{0.818} & \textbf{39.51}/\textbf{0.989} & - & - \\
\hline \multirow{2}{*}{ 3} 
& AirNet & 33.92/0.933 & 31.26/0.888 & 28.01/0.798 & 34.90/0.968 & 27.94/0.961 & - \\
& PromptIR & 33.98/0.933 & 31.31/0.888 & 28.06/0.799 & 36.37/0.972 & \textbf{30.58}/\textbf{0.974} & - \\
& \textit{Ours} & \textbf{34.59}/\textbf{0.941} & \textbf{31.83}/\textbf{0.900} & \textbf{28.46}/\textbf{0.814} & \textbf{37.50}/\textbf{0.980} & 29.20/0.972 & - \\
\hline \multirow{2}{*}{ 4} 
& AirNet & 33.89/0.932 & 31.21/0.887 & 27.97/0.795 & 34.70/0.964 & 27.41/0.956 & 26.36/0.799 \\
& PromptIR & 33.91/0.933 & 31.24/0.888 & 28.01/0.797 & 36.14/0.968 & \textbf{29.82}/0.969 & 27.16/0.820 \\
& \textit{Ours} & \textbf{34.58}/\textbf{0.941} & \textbf{31.83}/\textbf{0.900} & \textbf{28.46}/\textbf{0.813} & \textbf{37.35}/\textbf{0.980} & 28.93/\textbf{0.971} & \textbf{27.42}/\textbf{0.829} \\
\bottomrule
\end{tabular}}
\label{tab:multiple}
\end{table*}

\subsection{Results on various combined degradations}
In this section, we examine the impact of various combinations of degradation types on model performance, as detailed in Table \ref{tab:combined}. When we randomly select and combine two out of the four degradation types, we observe that denoising performance remains relatively stable, regardless of whether it is combined with deraining, dehazing, or deblurring. This stability is likely because the denoising task dominates the training process, benefiting from a larger dataset across three noise levels: $\sigma=15$, $\sigma=25$, and $\sigma=50$.

For the deraining task, performance is optimal when combined with denoising, compared to combinations with dehazing or deblurring. This is likely due to both deraining and denoising focusing on recovering high-frequency details, thus aligning their frequency optimization directions. Conversely, dehazing and deblurring aim to remove low-frequency content, and their performance is enhanced when combined with denoising, due to the larger training dataset.

Similar trends are observed when we randomly select and combine three out of the four degradation types, further supporting these findings.
\begin{table*}[t!]
\centering
\caption{\textbf{Results on various combined degradations.} Tasks can enhance each other when their degradation types (\eg, deraining and denoising) share similar frequency optimization directions. Conversely, when degradation tasks (\eg, deraining and dehazing) have conflicting optimization goals, a performance drop is observed. }
\resizebox{0.99 \linewidth}{!}{
\begin{tabular}{@{}lccccccccccccr@{}}
\toprule
\multicolumn{4}{c}{ Degradation } & \multicolumn{3}{c}{ Denoise } & Derain & Dehaze & Deblur \\
\cmidrule(lr){1-4} \cmidrule(lr){5-7} \cmidrule(lr){7-7} \cmidrule(lr){8-8} 
Noise&Rain&Haze&blur&BSD68 $(\sigma=15)$ & BSD68 $(\sigma=25)$ & BSD68 $(\sigma=50)$ & Rain100L & SOTS & GOPro \\
\hline
$\checkmark$&$\checkmark$&&&34.69/0.942 & 31.93/0.902 & 28.59/0.818 & 38.93/0.984 & - & - \\
\rowcolor{Gray}
$\checkmark$&&$\checkmark$&&34.66/0.942 & 31.91/0.902 & 28.56/0.818 & - & 29.01/0.972 & - \\
$\checkmark$&&&$\checkmark$&34.67/0.942 & 31.92/0.902 & 28.57/0.817 & - & - & 29.05/0.871 \\
\rowcolor{Gray}
&$\checkmark$&$\checkmark$&& - & - & - & 36.55/0.976 & 28.64/0.971 & - \\
&$\checkmark$&&$\checkmark$& - & - & - & 37.99/0.981 & - & 28.69/0.863 \\
\rowcolor{Gray}
&&$\checkmark$&$\checkmark$& - & - & - & - & 28.02/0.968 & 26.74/0.809 \\
\midrule
\midrule
$\checkmark$&$\checkmark$&$\checkmark$&& 34.59/0.941 & 31.83/0.900 & 28.46/0.814 & 37.50/0.980 & 29.20/0.972 & - \\
\rowcolor{Gray}
$\checkmark$&$\checkmark$&&$\checkmark$&34.65/0.942 & 31.89/0.901 & 28.54/0.816 & 38.72/0.984 & - & 28.99/0.870 \\
$\checkmark$&&$\checkmark$&$\checkmark$&34.62/0.942 & 31.87/0.901 & 28.52/0.815 & - & 28.65/0.970 & 28.23/0.851 \\
\rowcolor{Gray}
&$\checkmark$&$\checkmark$&$\checkmark$& - & - & - & 36.03/0.974 & 28.15/0.968 & 26.70/0.809 \\
\bottomrule
\end{tabular}}
\label{tab:combined}
\end{table*}

\begin{figure*}[htbp]
  \centering
  \subfloat[Denoising at different $\sigma$ values]{\includegraphics[width=0.5\textwidth]{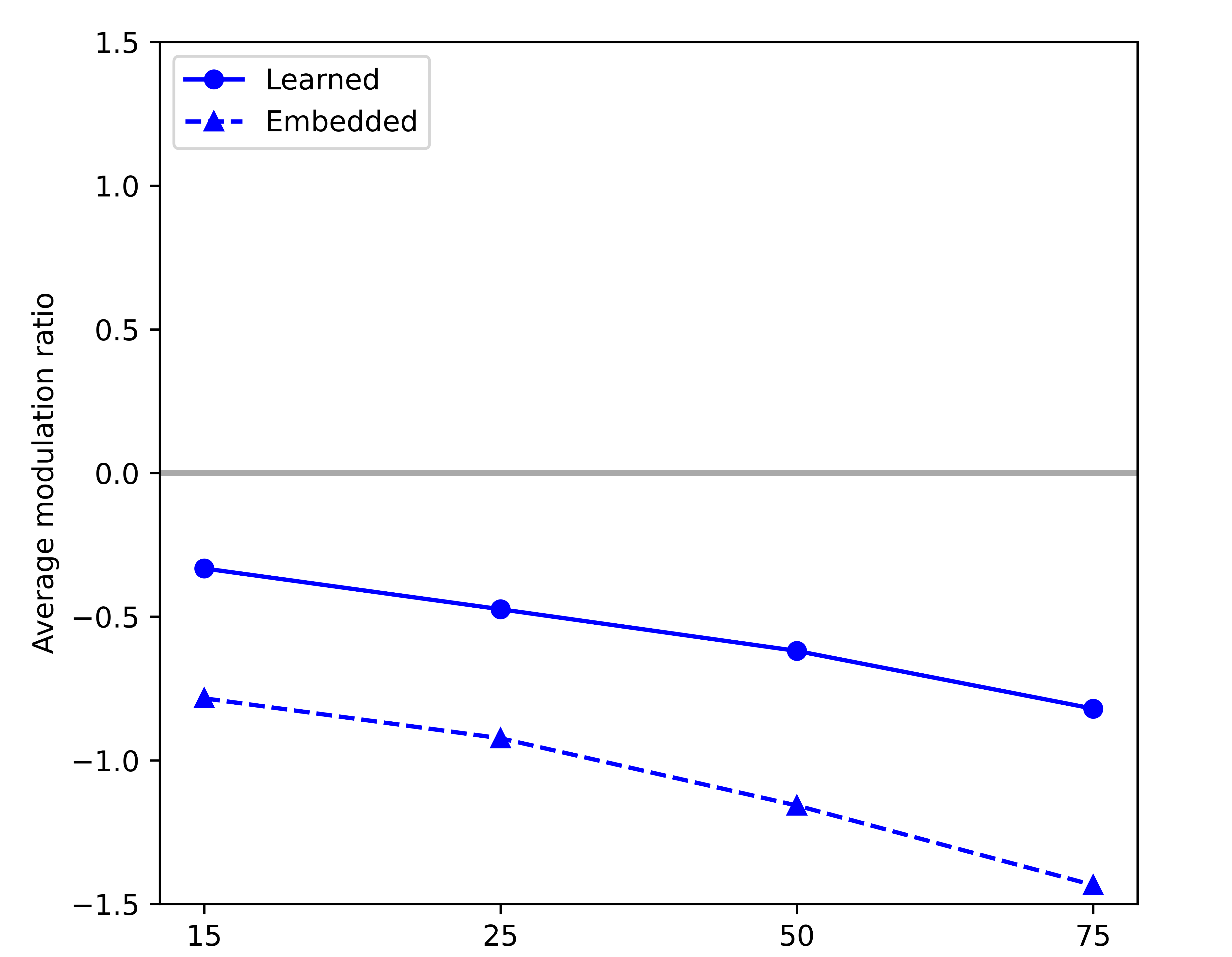}}
  \hfill
  \subfloat[4 D-types Image Restoration]{\includegraphics[width=0.5\textwidth]{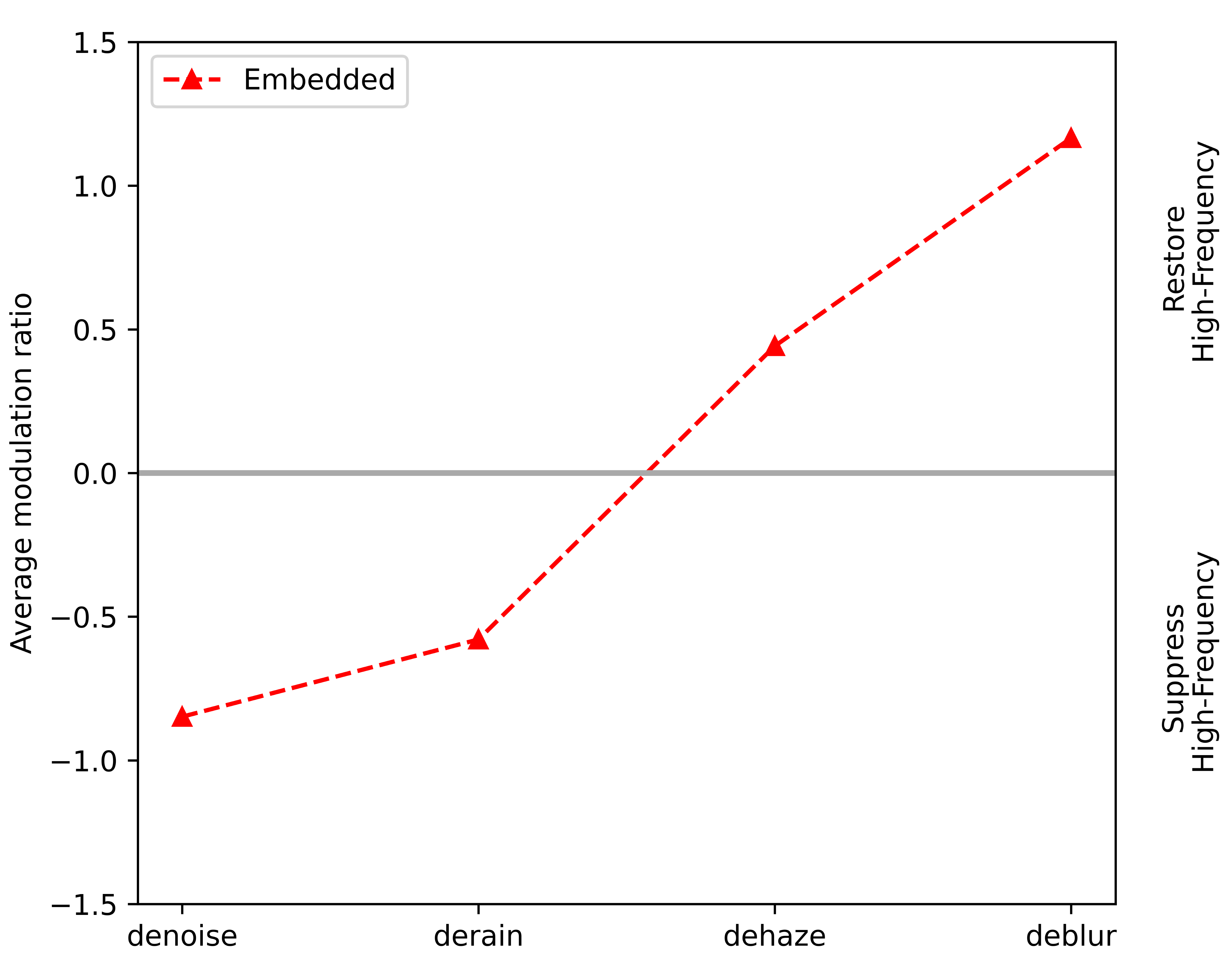}}
  \caption{\textbf{Average modulation ratio on different degradation type.} \textit{Learned}, directly learned average modulation ratios under One-by-one setting. \textit{Embedded}, average modulation ratios embedded from Dformer under All-in-one setting.}
  \label{fig:modulation}
\end{figure*}

\begin{figure*}[htbp]
  \centering
  \subfloat[PromptIR]{\includegraphics[width=0.33\textwidth]{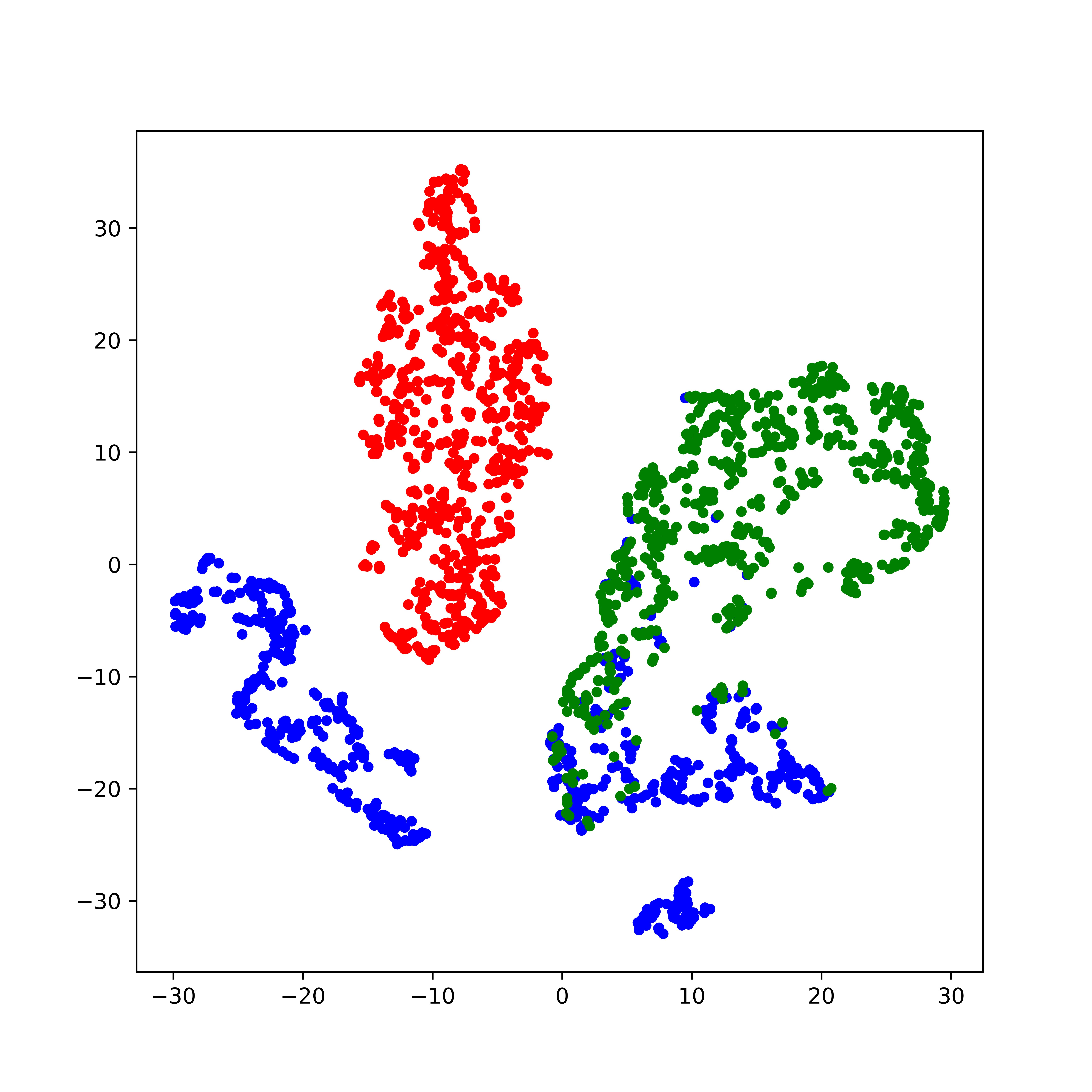}}
  \hfill
  \subfloat[Uformer]{\includegraphics[width=0.33\textwidth]{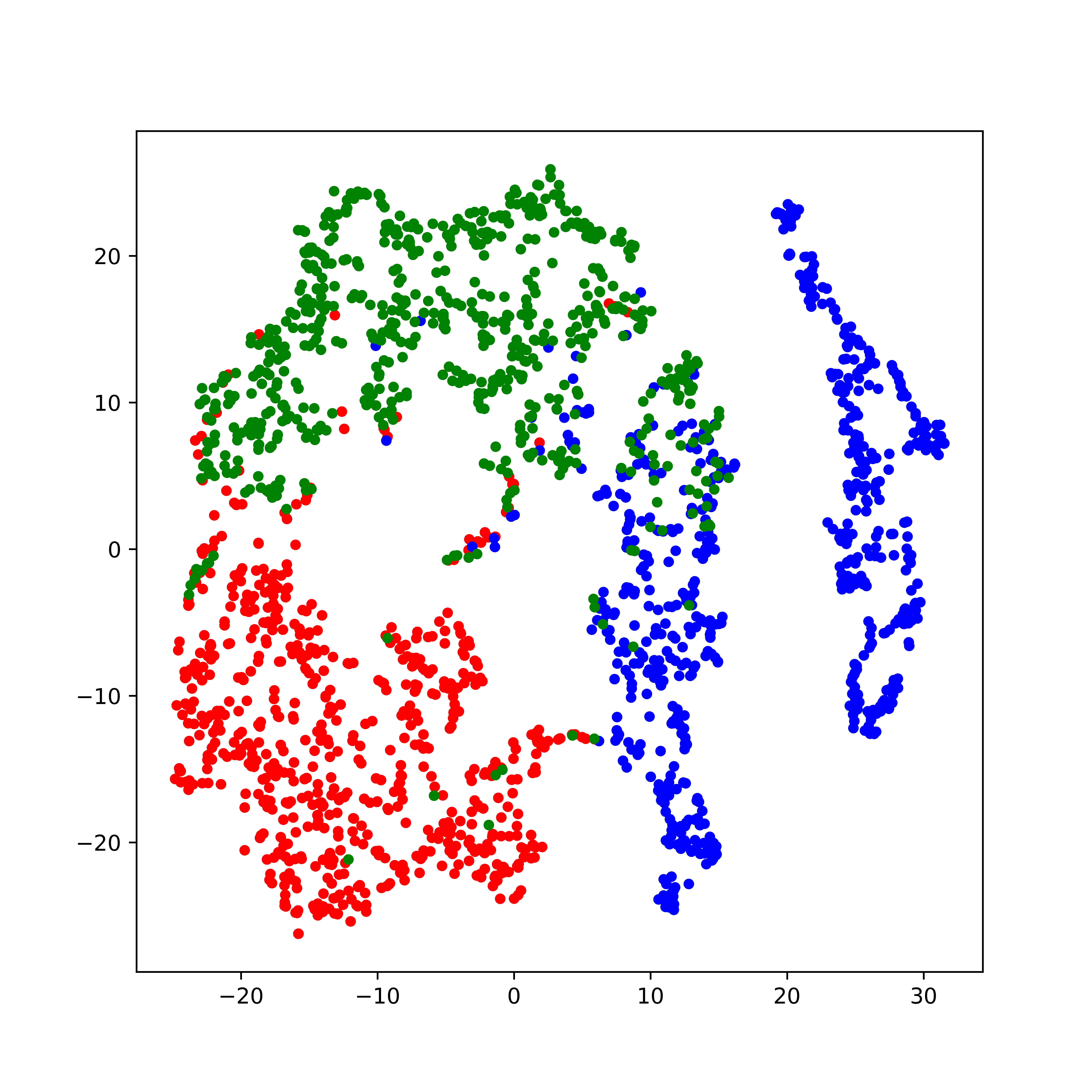}}
  \hfill
  \subfloat[\textit{Ours}]{\includegraphics[width=0.33\textwidth]{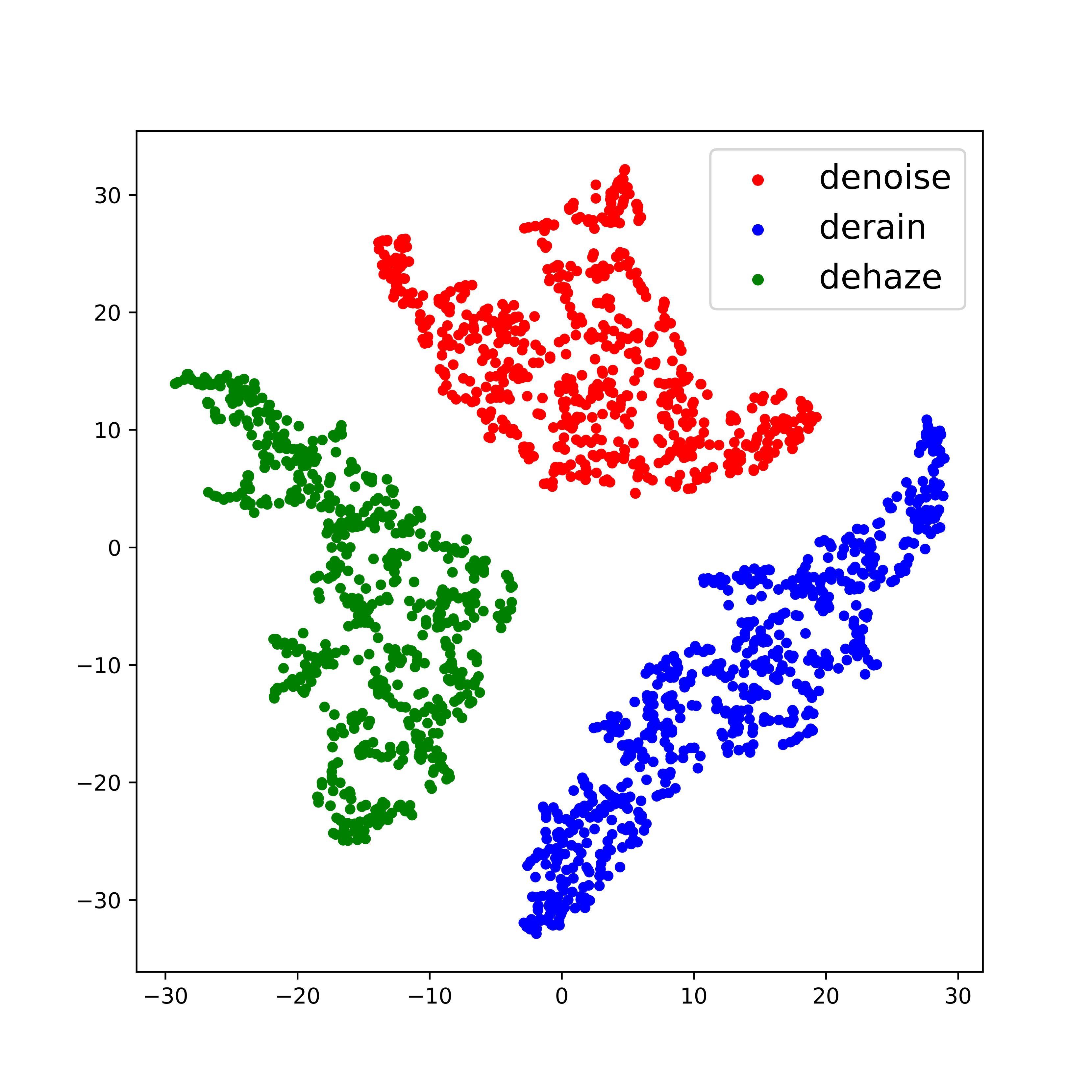}}
  \caption{\textbf{t-SNE visualization.} Without frequency modulation, Uformer \cite{Wang_2022_CVPR} fails to provide a discriminative capability superior to PromptIR\cite{potlapalli2023promptir}. However, with the incorporation of frequency modulation, our method effectively distinguish between different degradation types.}
  \label{fig:tsne}
\end{figure*}

\subsection{Analysis on Frequency Modulation}
In this section, we analyze whether the frequency modulation of the attention map captures the aforementioned frequency characteristics of various degradations.

First, we let the modulation ratios in Rformer be learnable, instead of embedding them from Dformer. Under the One-by-one setting with L=2, we train 4 models for the denoising task under Gaussian noise levels of $\sigma$=15,25,50,75, respectively, and visualize the average of the learned modulation ratios, obtaining the "Learned" entry in Fig. \ref{fig:modulation}(a). 
Since we do not modulate the lowest frequency in our method, i.e., the DC component, the average of the modulation parameters here is the average modulation ratio for high frequencies. From this, we find that even without using our Dformer to capture frequency characteristics, the model can still adapt to the noise level requirements by adjusting the proportion of different frequency bands in the attention map. For all four noise levels, the average high-frequency modulation ratio is less than 0, showing a trend of suppressing high frequencies. As the noise level increases, the high-frequency noise becomes larger, so the model needs to suppress high frequencies more, and the modulation ratio becomes smaller. Of course, although directly learnable modulation ratios can capture the frequency differences of different degradation levels, it is not suitable for the all-in-one setting because it requires pre-specifying the degradation type to obtain the corresponding modulation ratios.

Then, we replace the learnable modulation ratios with those embedded by Dformer and performe the same experiment under the all-in-one setting. Since the modulation ratios are no longer model parameters but are embedded from the input degraded image, we sample 100 images for each of the four noise levels and average the resulting average modulation ratios, obtaining the "Embedded" entry in Fig. \ref{fig:modulation}(a), finding that it still shows the same trend. Finally, we performe the same experiment on the 4 tasks All-in-one of denoising, deraining, dehazing, and deblurring, obtaining the "Embedded" entry in Fig. \ref{fig:modulation}(b), finding that denoising and deraining aim to suppress high-frequency noise, as the average modulation ratios is less than 0, while dehazing and deblurring are the opposite. In Fig. \ref{fig:frequency} and Fig. \ref{fig:modulation}, we can further find that even when comparing the dehazing and deblurring, which both restore high-frequency details, there are still some differences in frequency, that is, deblurring needs to restore high-frequency details more than dehazing.

To demonstrate that the proposed method not only captures frequency characteristics of various degradations but also effectively discriminates between different degradations, we perform t-SNE visualization on standard “noise-rain-haze” setting on the latent degradation representations obtained by PromptIR \cite{potlapalli2023promptir}, Uformer \cite{Wang_2022_CVPR} (i.e., the Rformer without frequency modulation), and our method. The results are shown in Fig. \ref{fig:tsne}. Without frequency modulation, Uformer \cite{Wang_2022_CVPR} fails to provide a discriminative capability superior to PromptIR\cite{potlapalli2023promptir}. However, with the incorporation of frequency modulation, our method effectively distinguish between different degradation types.

\subsection{Ablation Studies}
In this section, we conduct all ablation experiments in Table \ref{tab:ablation} on the standard “noise-rain-haze” setting to validate the components of the proposed network architecture. 

\textbf{Intra- \& Inter-Band attention.}
We conduct this ablation experiment to demonstrate the effectiveness of the proposed Intra-Band \& Inter-Band attention mechanism and the rationale behind the separation strategy of frequency bands. We replace the proposed attention mechanism with the origin window-based multi-head self-attention mechanism proposed by Swin Transformer \cite{liu2021swin}. As a result, a significant performance drop of 0.48 dB PSNR in dehazing task can be found in "w.o. intra-inter" in Table \ref{tab:ablation}, as the origin attention mechanism struggles to capture the interactions among different frequency bands.

\textbf{Frequency Decomposition.}
To illustrate the importance of frequency decomposition in the head layers of the Encoder to suppress the inherent bias of Transformer-like model towards different frequency information, we replaced the frequency decomposition layers in the Encoder with a simple MLP layer to achieve the same goal of dividing the features into $L$ sub-maps. In the "w.o. Decompose" column of Table \ref{tab:ablation}, we observed a certain degree of performance degradation across all 3 degradations. Specifically, deraining and dehazing experienced a decrease of 2 dB and 0.82 dB, respectively, in terms of PSNR. The significant performance drop indicates that without frequency decomposition, there is a pronounced negative interactions between different frequency bands, thereby impairing the performance of the Decoder when embedding degradation representations.

\begin{table*}[t!]
\centering
\caption{\textbf{Ablation staudy.} "w.o. intra-inter" means replacing the dual MSA with the regular MSA;"w.o. Decompose" means removing the initial frequency decomposition in the encoder and replacing the dual MSA with the regular MSA..}
\resizebox{0.9 \linewidth}{!}{
\begin{tabular}{@{}lccccccccccccr@{}}
\toprule
\multicolumn{2}{c}{ Frequency-aware} & \multicolumn{3}{c}{ Denoise } & Derain & Dehaze \\
\cmidrule(lr){3-5} \cmidrule(lr){6-6} \cmidrule(lr){7-7}  
Decompose&Intra- \& Inter-Band& BSD68 $(\sigma=15)$ & BSD68 $(\sigma=25)$ & BSD68 $(\sigma=50)$ & Rain100L & SOTS  \\
\hline
\rowcolor{Gray}
& & 34.59/0.942 & 31.84/0.900 & 28.48/0.814 & 37.42/0.980 & 28.17/0.968 \\
$\checkmark$ & & 34.59/0.941 & 31.83/0.900 & 28.48/0.813 & 37.45/0.980 & 28.72/0.971 \\
\rowcolor{Gray}
&$\checkmark$ & 34.44/0.940 & 31.71/0.898 & 28.35/0.809 & 35.50/0.970 & 28.38/0.968 \\
$\checkmark$ &$\checkmark$ & \textbf{34.59}/\textbf{0.941} & \textbf{31.83}/\textbf{0.900} & \textbf{28.46}/\textbf{0.814} & \textbf{37.50}/\textbf{0.980} & \textbf{29.20}/\textbf{0.972}\\
\bottomrule
\end{tabular}}
\label{tab:ablation}
\end{table*}

\begin{table*}[t!]
\centering
\caption{\textbf{Comparison to the state-of-the-art} on 5 tasks All-in-one setting. Our method consistently achieves state-of-the-art results.}
\resizebox{0.99 \linewidth}{!}{
\begin{tabular}{@{}lcccccccccccccr@{}}
\toprule
 \multirow{2}{*}{ Method }& \multirow{2}{*}{Params.}  & Dehaze & Derain & Denoise & Deblur & Low-light & \multirow{2}{*}{ Average } \\
\cmidrule(lr){3-3} \cmidrule(lr){4-4} \cmidrule(lr){5-5} \cmidrule(lr){6-6} \cmidrule(lr){7-7}
&& SOTS & Rain100L & BSD68 $(\sigma=25)$ & GOPro & LOL\\
\hline
DL\cite{8750830} &2M& 20.54/0.826 & 21.96/0.762 & 23.09/0.745 & 19.86/0.672 & 19.83/0.712 & 21.05/0.743\\
Transweather\cite{Valanarasu2021TransWeatherTR} &38M & 21.32/0.885 & 29.43/0.905 & 29.00/0.841 & 25.12/0.757 & 21.21/0.792 & 25.22/0.836\\
TAPE\cite{liu2022tape} &1M& 22.16/0.861 & 29.67/0.904 & 30.18/0.855 & 24.47/0.763 & 18.97/0.621 & 25.09/0.801\\
AirNet\cite{li2022all} &9M& 21.04/0.884 & 32.98/0.951 & 30.91/0.882 & 24.35/0.781 & 18.18/0.735 & 25.49/0.846\\
IDR\cite{zhang2023ingredient} &15M& 25.24/0.943 & 35.63/0.965 & 31.60/0.887 & 27.87/0.846 & 21.34/0.826 & 28.34/0.893\\
PromptIR\cite{potlapalli2023promptir} &36M& 26.54/0.949 & 36.37/0.970 & 31.47/0.886 & 28.71/0.881 & 22.68/0.832 & 29.15/0.904 \\
Gridformer\cite{wang2024gridformer} &34M& 26.79/0.951 & 36.61/0.971 & 31.45/0.885 & \textbf{29.22}/\textbf{0.884} & 22.59/0.831 & 29.33/0.904\\
\hline 
\rowcolor{Gray}
\textit{Ours}&32M&\textbf{27.04}/\textbf{0.961}&\textbf{37.51}/\textbf{0.978}&\textbf{31.77}/\textbf{0.893}&29.13/\textbf{0.884}&\textbf{23.02}/\textbf{0.840}&\textbf{29.69}/\textbf{0.911} \\
\bottomrule
\end{tabular}}
\label{tab:sota-5}
\end{table*}

\begin{table*}[t!]
\centering
\caption{\textbf{Comparison on real-world denoising. } Our method achieves
state-of-the-art performance, validating its effectiveness in
handling complex real-world scenarios.}
\resizebox{0.99 \linewidth}{!}{
\begin{tabular}{@{}lccccccccc@{}}
\toprule
Method & MIRNet\cite{Zamir2020MIRNet} & DeamNet\cite{Ren_2021_CVPR} & MPRNet\cite{Zamir2021MPRNet} & DAGL\cite{mou2021gatir} & Uformer\cite{Wang_2022_CVPR} & Restormer\cite{Zamir2021Restormer} & SSAMAN\cite{zafar2024single}& NAFNet\cite{chu2022nafssr} & \textit{Ours} \\
\hline
SIDD&39.72/0.959 &39.47/0.957 &39.71/0.958 &38.94/0.953 &39.77/0.959 & 40.02/0.960 & 40.08/0.962 & 40.30/0.961 &\textbf{40.38}/\textbf{0.971}\\
     
\bottomrule
\end{tabular}}
\label{tab:sidd}
\end{table*}

\subsection{Generalization analysis}
\textbf{Performance on 5 tasks All-in-one setting.}
Informed by recent studies, including Gridformer \cite{wang2024gridformer} and InstructIR \cite{conde2024high}, we extended the existing four degradation types by incorporating the LOL dataset \cite{wei2018deep} for low-light degradation to further validate the efficacy of the proposed method. A comprehensive experiment encompassing all five degradation types was then conducted. Furthermore, we compared our method against a broader range of literature focusing on the same task setting. The experimental results are detailed in Table \ref{tab:sota-5}. As demonstrated, our method achieves a performance gain of 0.36 dB over the current state-of-the-art method.

\textbf{Performance on real-world denoising.}
To further demonstrate that our method is not limited to synthetically injected noise, we followed the experimental setup of Restormer\cite{Zamir2021Restormer} and conducted experiments on the real-world image denoising dataset SIDD \cite{SIDD_2018_CVPR}. Specifically, without leveraging any external data, we trained and tested our model directly on the SIDD dataset. As shown in Table \ref{tab:sidd}, our method achieves state-of-the-art performance, validating its effectiveness in handling complex real-world scenarios.

\textbf{Performance on spatially variant degradation.}
We analyze the performance of the proposed method under spatially variant degradation, aiming to highlight its enhanced capability in restoring spatial heterogeneity. Following the experimental setup of AirNet \cite{li2022all}, we partition each clean image of the BSD68 \cite{martin2001database} dataset into four regions. Subsequently, Gaussian noise with $\sigma \in \{0,15,25,50\}$ is injected into each region individually to create a new test set. We then assess the model trained solely on the standard denoising task using this new test set. As shown in Table \ref{tab:generalization-1}, our method outperforms both AirNet \cite{li2022all} and PromptIR \cite{potlapalli2023promptir}, achieving a PSNR improvements of 0.34 dB over AirNet and 0.11 dB over PromptIR.
\begin{table}[t!]
\centering
\caption{\textbf{Performance on spatially variant degradation.} Under spatially variant degradation, our method showcases superior denoising performance compared to existing methods.}
\resizebox{0.6 \linewidth}{!}{
\begin{tabular}{@{}lccccccccccccr@{}}
\toprule
\multirow{2}{*}{ Method } & Denoise \\
\cmidrule(lr){2-2}
& BSD68 $(\sigma \in \{0,15,25,50\})$ \\
\hline 
AirNet & 31.42/0.892 \\
PromptIR & 31.65/0.899 \\
\textit{Ours} & \textbf{31.76/0.902} \\
\bottomrule
\end{tabular}}
\label{tab:generalization-1}
\end{table}

\begin{table}[t!]
\centering
\caption{\textbf{Generalization to unseen degradation levels.} Our method achieves superior generalization performance over the existing AirNet and PromptIR.}
\resizebox{0.8 \linewidth}{!}{
\begin{tabular}{@{}lccccccccccccr@{}}
\toprule
\multirow{2}{*}{ Method } & \multicolumn{2}{c}{Denoise} \\
\cmidrule(lr){2-3}
& BSD68 $(\sigma \in [15,25])$ & BSD68 $(\sigma \in [25,50])$ \\
\hline 
AirNet & 31.80/0.887 & 28.30/0.782 \\
PromptIR & 32.34/0.908 & 29.18/0.830 \\
\textit{Ours} & \textbf{33.13/0.918} & \textbf{29.34/0.832} \\
\bottomrule
\end{tabular}}
\label{tab:generalization-2}
\end{table}

\begin{table}[t!]
\centering
\caption{\textbf{Generalization to mixed degradations.} Even with mixed degradations, our method maintains superior generalization performance.}
\resizebox{0.76 \linewidth}{!}{
\begin{tabular}{@{}lccccccccccccr@{}}
\toprule
\multirow{2}{*}{ Method } & Denoise + Derain & Denoise + Deblur\\
\cmidrule(lr){2-2} \cmidrule(lr){3-3}
& Rain100L $(\sigma = 25)$ & GoPro $(\sigma = 25)$ \\
\hline 
AirNet & 29.17/0.858 & 22.61/0.667\\
PromptIR & 29.41/0.857 & 22.50/0.665\\
\textit{Ours} & \textbf{30.02}/\textbf{0.862} & \textbf{25.03}/\textbf{0.728}\\
\bottomrule
\end{tabular}}
\label{tab:mixed}
\end{table}

\textbf{Generalization to unseen degradation levels.}
To analyze the generalization capability of our model for unseen degradation levels, we evaluate it on the BSD68 \cite{martin2001database} test set. Specifically, our model, trained solely on $\sigma \in \{15,25,50\}$, is tested with randomly sampled values from the ranges $\sigma \in [15,25]$ and $\sigma \in [25,50]$. The results shown in Table \ref{tab:generalization-2} highlight the superior generalization performance of our model over AirNet and PromptIR.

\textbf{Generalization to mixed degradations.}
To further analyze the generalization capability, we conducted experiments by mixing multiple degradations within the same image. Specifically, we injected Gaussian noise with a standard deviation of $\sigma$=25 to the degraded images from the rain removal dataset Rain100L \cite{yang2019joint} and the blur removal dataset GoPro \cite{nah2017deep}, constructing a test set to assess the models trained under the 4 tasks All-in-one setting. The results, presented in Table \ref{tab:mixed}, demonstrate that our model exhibits superior generalization to mixed degradations.


\textbf{The effect of the number of frequency bands.} 
\label{sec:analysis-frequency bands}
Finally, we examine the impact of the number of frequency bands, denoted as $L$, on both efficiency and performance. Specifically, we compare the performance between $L=2$ and $L=3$ under the standard “noise-rain-haze" setting, as detailed in Table \ref{tab:bands}. The results show that increasing the number of frequency bands from $L=2$ to $L=3$ improves restoration performance across all three tasks. This aligns with our expectations, as a higher value of $L$ enables the model to more finely distinguish between different degradation types at various frequencies, enhancing overall performance. However, the number of tokens in Intra- and Inter-Band attention scales with $L$, leading to attention maps and time complexity increasing proportionally. For example, the training time per epoch for Dformer is 70 seconds for $L=2$ and 90 seconds for $L=3$, with a corresponding increase in FLOPs from 213.35G to 231.10G. To balance efficiency and performance, we have chosen to use $L=2$ as the default value for all experiments.
\begin{table*}[t!]
\centering
\caption{\textbf{The effect of frequency decomposition } on three degradation types. The restoration performance for all three tasks boosts when the number of frequency bands is increased from $L=2$ to $L=3$, while efficiency decreases}
\resizebox{0.99 \linewidth}{!}{
\begin{tabular}{@{}lccccccccccccr@{}}
\toprule
\multirow{2}{*}{ Method } & \multicolumn{3}{c}{ Denoise } & Derain & Dehaze & \multirow{2}{*}{FLOPs}&\multirow{2}{*}{ Training time of Dformer} \\
\cmidrule(lr){2-4} \cmidrule(lr){5-5} \cmidrule(lr){6-6}
& BSD68 $(\sigma=15)$ & BSD68 $(\sigma=25)$ & BSD68 $(\sigma=50)$ & Rain100L & SOTS \\
\hline
L=2 & 34.59/0.941 & 31.83/0.900 & 28.46/0.814 & 37.50/0.980 & 29.20/0.972 & \textbf{213.35G} & \textbf{70s/epoch} \\
L=3 & \textbf{34.61}/\textbf{0.944} & \textbf{31.92}/\textbf{0.902} & \textbf{28.54}/\textbf{0.816} & \textbf{37.88}/\textbf{0.982} & \textbf{29.33}/\textbf{0.974} & 231.10G & 90s/epoch \\
\bottomrule
\end{tabular}}
\label{tab:bands}
\end{table*}

\section{Conclusion}
This work presents an all-in-one image restoration model leveraging advanced vision transformers, inspired by the fact that various degradations uniquely impact image content across different frequency bands. The model consists of two primary components: the frequency-aware Degradation Prior Learning Transformer (Dformer) and the Degradation-Adaptive Restoration Transformer (Rformer). The Dformer captures degradation representations by using an input frequency decomposition module and frequency-aware Swin Transformer blocks. Guided by these learned representations, the Rformer utilizes a degradation-adaptive self-attention module to selectively focus on the most affected frequency components for restoration. Extensive experimental results demonstrate the superiority of our approach over existing methods in five key restoration tasks: denoising, deraining, dehazing, deblurring, and low-light enhancement. Furthermore, our method excels in handling real-world degradations, spatially variant degradations, and previously unseen degradation levels. These findings underscore the potential of our frequency-based perspective and advanced transformer design to significantly advance the field of image restoration.


\bibliographystyle{IEEEtran}
\bibliography{ref}

\begin{thebibliography}{10}
\providecommand{\url}[1]{#1}
\csname url@samestyle\endcsname
\providecommand{\newblock}{\relax}
\providecommand{\bibinfo}[2]{#2}
\providecommand{\BIBentrySTDinterwordspacing}{\spaceskip=0pt\relax}
\providecommand{\BIBentryALTinterwordstretchfactor}{4}
\providecommand{\BIBentryALTinterwordspacing}{\spaceskip=\fontdimen2\font plus
\BIBentryALTinterwordstretchfactor\fontdimen3\font minus \fontdimen4\font\relax}
\providecommand{\BIBforeignlanguage}[2]{{%
\expandafter\ifx\csname l@#1\endcsname\relax
\typeout{** WARNING: IEEEtran.bst: No hyphenation pattern has been}%
\typeout{** loaded for the language `#1'. Using the pattern for}%
\typeout{** the default language instead.}%
\else
\language=\csname l@#1\endcsname
\fi
#2}}
\providecommand{\BIBdecl}{\relax}
\BIBdecl

\bibitem{zhang2017beyond}
K.~Zhang, W.~Zuo, Y.~Chen, D.~Meng, and L.~Zhang, ``Beyond a {Gaussian} denoiser: Residual learning of deep {CNN} for image denoising,'' \emph{IEEE Transactions on Image Processing}, vol.~26, no.~7, pp. 3142--3155, 2017.

\bibitem{zhang2018ffdnet}
K.~Zhang, W.~Zuo, and L.~Zhang, ``Ffdnet: Toward a fast and flexible solution for {CNN} based image denoising,'' \emph{IEEE Transactions on Image Processing}, 2018.

\bibitem{shi2021unsharp}
Z.~Shi, Y.~Chen, E.~Gavves, P.~Mettes, and C.~G. Snoek, ``Unsharp mask guided filtering,'' \emph{IEEE Transactions on Image Processing}, vol.~30, pp. 7472--7485, 2021.

\bibitem{shi2022measuring}
Z.~Shi, P.~Mettes, S.~Maji, and C.~G. Snoek, ``On measuring and controlling the spectral bias of the deep image prior,'' \emph{International Journal of Computer Vision}, vol. 130, no.~4, pp. 885--908, 2022.

\bibitem{zhang2018density}
H.~Zhang and V.~M. Patel, ``Density-aware single image de-raining using a multi-stream dense network,'' in \emph{CVPR}, 2018.

\bibitem{chen2021robust}
C.~Chen and H.~Li, ``Robust representation learning with feedback for single image deraining,'' in \emph{CVPR}, 2021.

\bibitem{yang2020single}
W.~Yang, R.~T. Tan, S.~Wang, Y.~Fang, and J.~Liu, ``Single image deraining: From model-based to data-driven and beyond,'' \emph{IEEE Transactions on pattern analysis and machine intelligence}, vol.~43, no.~11, pp. 4059--4077, 2020.

\bibitem{zheng2020single}
Y.~Zheng, X.~Yu, M.~Liu, and S.~Zhang, ``Single-image deraining via recurrent residual multiscale networks,'' \emph{IEEE transactions on neural networks and learning systems}, vol.~33, no.~3, pp. 1310--1323, 2020.

\bibitem{hu2021pyramid}
X.~Hu, W.~Ren, K.~Yu, K.~Zhang, X.~Cao, W.~Liu, and B.~Menze, ``Pyramid architecture search for real-time image deblurring,'' in \emph{ICCV}, 2021.

\bibitem{kupyn2018deblurgan}
O.~Kupyn, V.~Budzan, M.~Mykhailych, D.~Mishkin, and J.~Matas, ``Deblurgan: Blind motion deblurring using conditional adversarial networks,'' in \emph{CVPR}, 2018.

\bibitem{rim2022realistic}
J.~Rim, G.~Kim, J.~Kim, J.~Lee, S.~Lee, and S.~Cho, ``Realistic blur synthesis for learning image deblurring,'' in \emph{ECCV}, 2022.

\bibitem{zhang2022blind}
J.~Zhang and W.~Zhai, ``Blind attention geometric restraint neural network for single image dynamic/defocus deblurring,'' \emph{IEEE Transactions on Neural Networks and Learning Systems}, vol.~34, no.~11, pp. 8404--8417, 2022.

\bibitem{chen2020pre}
H.~Chen, Y.~Wang, T.~Guo, C.~Xu, Y.~Deng, Z.~Liu, S.~Ma, C.~Xu, C.~Xu, and W.~Gao, ``Pre-trained image processing transformer,'' in \emph{CVPR}, 2021.

\bibitem{9157460}
R.~Li, R.~T. Tan, and L.-F. Cheong, ``All in one bad weather removal using architectural search,'' in \emph{CVPR}, 2020.

\bibitem{li2022all}
B.~Li, X.~Liu, P.~Hu, Z.~Wu, J.~Lv, and X.~Peng, ``All-in-one image restoration for unknown corruption,'' in \emph{CVPR}, 2022.

\bibitem{zhang2023all}
C.~Zhang, Y.~Zhu, Q.~Yan, J.~Sun, and Y.~Zhang, ``All-in-one multi-degradation image restoration network via hierarchical degradation representation,'' in \emph{ACM MM}, 2023.

\bibitem{potlapalli2023promptir}
V.~Potlapalli, S.~W. Zamir, S.~Khan, and F.~Khan, ``Promptir: Prompting for all-in-one image restoration,'' in \emph{NeurIPS}, 2023.

\bibitem{ai2024multimodal}
Y.~Ai, H.~Huang, X.~Zhou, J.~Wang, and R.~He, ``Multimodal prompt perceiver: Empower adaptiveness generalizability and fidelity for all-in-one image restoration,'' in \emph{CVPR}, 2024.

\bibitem{conde2024high}
M.~V. Conde, G.~Geigle, and R.~Timofte, ``Instructir: High-quality image restoration following human instructions,'' in \emph{ECCV}, 2024.

\bibitem{10204770}
D.~Park, B.~H. Lee, and S.~Y. Chun, ``All-in-one image restoration for unknown degradations using adaptive discriminative filters for specific degradations,'' in \emph{CVPR}, 2023.

\bibitem{jiang2023autodir}
Y.~Jiang, Z.~Zhang, T.~Xue, and J.~Gu, ``Autodir: Automatic all-in-one image restoration with latent diffusion,'' in \emph{ECCV}, 2024.

\bibitem{10204072}
J.~Zhang, J.~Huang, M.~Yao, Z.~Yang, H.~Yu, M.~Zhou, and F.~Zhao, ``Ingredient-oriented multi-degradation learning for image restoration,'' in \emph{CVPR}, 2023.

\bibitem{ren2016image}
W.~Ren, X.~Cao, J.~Pan, X.~Guo, W.~Zuo, and M.-H. Yang, ``Image deblurring via enhanced low-rank prior,'' \emph{IEEE Transactions on Image Processing}, vol.~25, no.~7, pp. 3426--3437, 2016.

\bibitem{park2023all}
D.~Park, B.~H. Lee, and S.~Y. Chun, ``All-in-one image restoration for unknown degradations using adaptive discriminative filters for specific degradations,'' in \emph{CVPR}, 2023.

\bibitem{Dabov2007ColorID}
K.~Dabov, A.~Foi, V.~Katkovnik, and K.~O. Egiazarian, ``Color image denoising via sparse 3d collaborative filtering with grouping constraint in luminance-chrominance space,'' in \emph{ICIP}, 2007.

\bibitem{tian2020image}
C.~Tian, Y.~Xu, and W.~Zuo, ``Image denoising using deep cnn with batch renormalization,'' \emph{Neural Networks}, vol. 121, pp. 461--473, 2020.

\bibitem{cai2016dehazenet}
B.~Cai, X.~Xu, K.~Jia, C.~Qing, and D.~Tao, ``Dehazenet: An end-to-end system for single image haze removal,'' \emph{IEEE Transactions on Image Processing}, vol.~25, no.~11, pp. 5187--5198, 2016.

\bibitem{dong2021fdgan}
Y.~Dong, Y.~Liu, H.~Zhang, S.~Chen, and Y.~Qiao, ``Fd-gan: Generative adversarial networks with fusion-discriminator for single image dehazing,'' in \emph{AAAI}, 2020.

\bibitem{5567108}
K.~He, J.~Sun, and X.~Tang, ``Single image haze removal using dark channel prior,'' \emph{IEEE Transactions on Pattern Analysis and Machine Intelligence}, vol.~33, no.~12, pp. 2341--2353, 2011.

\bibitem{zhou2022fsad}
Y.~Zhou, Z.~Chen, P.~Li, H.~Song, C.~P. Chen, and B.~Sheng, ``Fsad-net: feedback spatial attention dehazing network,'' \emph{IEEE transactions on neural networks and learning systems}, vol.~34, no.~10, pp. 7719--7733, 2022.

\bibitem{Wei_2019_CVPR}
W.~Wei, D.~Meng, Q.~Zhao, Z.~Xu, and Y.~Wu, ``Semi-supervised transfer learning for image rain removal,'' in \emph{CVPR}, 2019.

\bibitem{ren2019progressive}
D.~Ren, W.~Zuo, Q.~Hu, P.~Zhu, and D.~Meng, ``Progressive image deraining networks: A better and simpler baseline,'' in \emph{CVPR}, 2019.

\bibitem{Kui_2020_CVPR}
K.~Jiang, Z.~Wang, P.~Yi, C.~Chen, B.~Huang, Y.~Luo, J.~Ma, and J.~Jiang, ``Multi-scale progressive fusion network for single image deraining,'' in \emph{CVPR}, 2020.

\bibitem{8099669}
X.~Fu, J.~Huang, D.~Zeng, Y.~Huang, X.~Ding, and J.~Paisley, ``Removing rain from single images via a deep detail network,'' in \emph{CVPR}, 2017.

\bibitem{Cho2021RethinkingCA}
S.-J. Cho, S.~Ji, J.-P. Hong, S.~Jung, and S.-J. Ko, ``Rethinking coarse-to-fine approach in single image deblurring,'' in \emph{ICCV}, 2021.

\bibitem{7780549}
J.~Pan, D.~Sun, H.~Pfister, and M.-H. Yang, ``Blind image deblurring using dark channel prior,'' in \emph{CVPR}, 2016.

\bibitem{Cui_Tao_Ren_Knoll_2023}
Y.~Cui, Y.~Tao, W.~Ren, and A.~Knoll, ``Dual-domain attention for image deblurring,'' in \emph{AAAI}, 2023.

\bibitem{liang2024image}
P.~Liang, J.~Jiang, X.~Liu, and J.~Ma, ``Image deblurring by exploring in-depth properties of transformer,'' \emph{IEEE Transactions on Neural Networks and Learning Systems}, 2024.

\bibitem{NIPS2017_3f5ee243}
A.~Vaswani, N.~Shazeer, N.~Parmar, J.~Uszkoreit, L.~Jones, A.~N. Gomez, L.~u. Kaiser, and I.~Polosukhin, ``Attention is all you need,'' in \emph{NeurIPS}, 2017.

\bibitem{Tsai2022Stripformer}
F.-J. Tsai, Y.-T. Peng, Y.-Y. Lin, C.-C. Tsai, and C.-W. Lin, ``Stripformer: Strip transformer for fast image deblurring,'' in \emph{ECCV}, 2022.

\bibitem{liang2021swinir}
J.~Liang, J.~Cao, G.~Sun, K.~Zhang, L.~Van~Gool, and R.~Timofte, ``Swinir: Image restoration using swin transformer,'' in \emph{ICCV}, 2021.

\bibitem{Zamir2021Restormer}
S.~W. Zamir, A.~Arora, S.~Khan, M.~Hayat, F.~S. Khan, and M.-H. Yang, ``Restormer: Efficient transformer for high-resolution image restoration,'' in \emph{CVPR}, 2022.

\bibitem{Wang_2022_CVPR}
Z.~Wang, X.~Cun, J.~Bao, W.~Zhou, J.~Liu, and H.~Li, ``Uformer: A general u-shaped transformer for image restoration,'' in \emph{CVPR}, 2022.

\bibitem{si2022inception}
C.~Si, W.~Yu, P.~Zhou, Y.~Zhou, X.~Wang, and S.~YAN, ``Inception transformer,'' in \emph{NeurIPS}, 2022.

\bibitem{NEURIPS2022_a37fea8e}
Z.~Chen, Y.~Zhang, J.~Gu, y.~zhang, L.~Kong, and X.~Yuan, ``Cross aggregation transformer for image restoration,'' in \emph{NeurIPS}, 2022.

\bibitem{zhang2023accurate}
J.~Zhang, Y.~Zhang, J.~Gu, Y.~Zhang, L.~Kong, and X.~Yuan, ``Accurate image restoration with attention retractable transformer,'' in \emph{ICLR}, 2023.

\bibitem{RonnebergerFB15}
O.~Ronneberger, P.~Fischer, and T.~Brox, ``U-net: Convolutional networks for biomedical image segmentation,'' in \emph{MICCAI}, 2015.

\bibitem{liu2021swin}
Z.~Liu, Y.~Lin, Y.~Cao, H.~Hu, Y.~Wei, Z.~Zhang, S.~Lin, and B.~Guo, ``Swin transformer: Hierarchical vision transformer using shifted windows,'' in \emph{ICCV}, 2021.

\bibitem{NEURIPS2020_c6e81542}
Y.~Gou, B.~Li, Z.~Liu, S.~Yang, and X.~Peng, ``Clearer: Multi-scale neural architecture search for image restoration,'' in \emph{NeurIPS}, 2020.

\bibitem{liu2018non}
D.~Liu, B.~Wen, Y.~Fan, C.~C. Loy, and T.~S. Huang, ``Non-local recurrent network for image restoration,'' in \emph{NeurIPS}, 2018.

\bibitem{Tai-MemNet-2017}
Y.~Tai, J.~Yang, X.~Liu, and C.~Xu, ``Memnet: A persistent memory network for image restoration,'' in \emph{ICCV}, 2017.

\bibitem{Zamir2021MPRNet}
S.~W. Zamir, A.~Arora, S.~Khan, M.~Hayat, F.~S. Khan, M.-H. Yang, and L.~Shao, ``Multi-stage progressive image restoration,'' in \emph{CVPR}, 2021.

\bibitem{8099783}
K.~Zhang, W.~Zuo, S.~Gu, and L.~Zhang, ``Learning deep cnn denoiser prior for image restoration,'' in \emph{CVPR}, 2017.

\bibitem{Wei_2021_CVPR}
Y.~Wei, S.~Gu, Y.~Li, R.~Timofte, L.~Jin, and H.~Song, ``Unsupervised real-world image super resolution via domain-distance aware training,'' in \emph{CVPR}, 2021.

\bibitem{radford2021learning}
A.~Radford, J.~W. Kim, C.~Hallacy, A.~Ramesh, G.~Goh, S.~Agarwal, G.~Sastry, A.~Askell, P.~Mishkin, J.~Clark \emph{et~al.}, ``Learning transferable visual models from natural language supervision,'' in \emph{ICML}, 2021.

\bibitem{luo2023controlling}
Z.~Luo, F.~K. Gustafsson, Z.~Zhao, J.~Sj{\"o}lund, and T.~B. Sch{\"o}n, ``Controlling vision-language models for universal image restoration,'' in \emph{ICLR}, 2024.

\bibitem{yang2024language}
H.~Yang, L.~Pan, Y.~Yang, and W.~Liang, ``Language-driven all-in-one adverse weather removal,'' in \emph{CVPR}, 2024.

\bibitem{liu2023unifying}
Y.~Liu, X.~Chen, X.~Ma, X.~Wang, J.~Zhou, Y.~Qiao, and C.~Dong, ``Unifying image processing as visual prompting question answering,'' in \emph{ICML}, 2023.

\bibitem{xu2020learning}
K.~Xu, M.~Qin, F.~Sun, Y.~Wang, Y.-K. Chen, and F.~Ren, ``Learning in the frequency domain,'' in \emph{CVPR}, 2020.

\bibitem{8803391}
H.-H. Yang and Y.~Fu, ``Wavelet u-net and the chromatic adaptation transform for single image dehazing,'' in \emph{ICIP}, 2019.

\bibitem{DeepRFT2021}
X.~Mao, Y.~Liu, F.~Liu, Q.~Li, W.~Shen, and Y.~Wang, ``Intriguing findings of frequency selection for image deblurring,'' in \emph{AAAI}, vol.~37, no.~2, 2023, pp. 1905--1913.

\bibitem{cui2023selective}
Y.~Cui, Y.~Tao, Z.~Bing, W.~Ren, X.~Gao, X.~Cao, K.~Huang, and A.~Knoll, ``Selective frequency network for image restoration,'' in \emph{ICLR}, 2023.

\bibitem{jiang2021focal}
L.~Jiang, B.~Dai, W.~Wu, and C.~C. Loy, ``Focal frequency loss for image reconstruction and synthesis,'' in \emph{ICCV}, 2021.

\bibitem{8578797}
N.~Kwak, J.~Yoo, and S.-h. Lee, ``Image restoration by estimating frequency distribution of local patches,'' in \emph{CVPR}, 2018.

\bibitem{hsu2023wavelet}
W.-Y. Hsu and P.-W. Jian, ``Wavelet pyramid recurrent structure-preserving attention network for single image super-resolution,'' \emph{IEEE Transactions on Neural Networks and Learning Systems}, 2023.

\bibitem{park2022vision}
N.~Park and S.~Kim, ``How do vision transformers work?'' in \emph{ICLR}, 2022.

\bibitem{wang2022antioversmoothing}
P.~Wang, W.~Zheng, T.~Chen, and Z.~Wang, ``Anti-oversmoothing in deep vision transformers via the fourier domain analysis: From theory to practice,'' in \emph{ICLR}, 2022.

\bibitem{martin2001database}
D.~Martin, C.~Fowlkes, D.~Tal, and J.~Malik, ``A database of human segmented natural images and its application to evaluating segmentation algorithms and measuring ecological statistics,'' in \emph{ICCV}, 2001.

\bibitem{ma2016waterloo}
K.~Ma, Z.~Duanmu, Q.~Wu, Z.~Wang, H.~Yong, H.~Li, and L.~Zhang, ``Waterloo exploration database: New challenges for image quality assessment models,'' \emph{IEEE Transactions on Image Processing}, vol.~26, no.~2, pp. 1004--1016, 2016.

\bibitem{huang2015single}
J.-B. Huang, A.~Singh, and N.~Ahuja, ``Single image super-resolution from transformed self-exemplars,'' in \emph{CVPR}, 2015.

\bibitem{yang2019joint}
W.~Yang, R.~T. Tan, J.~Feng, Z.~Guo, S.~Yan, and J.~Liu, ``Joint rain detection and removal from a single image with contextualized deep networks,'' \emph{IEEE transactions on pattern analysis and machine intelligence}, vol.~42, no.~6, pp. 1377--1393, 2019.

\bibitem{li2018benchmarking}
B.~Li, W.~Ren, D.~Fu, D.~Tao, D.~Feng, W.~Zeng, and Z.~Wang, ``Benchmarking single-image dehazing and beyond,'' \emph{IEEE Transactions on Image Processing}, vol.~28, no.~1, pp. 492--505, 2018.

\bibitem{nah2017deep}
S.~Nah, T.~Hyun~Kim, and K.~Mu~Lee, ``Deep multi-scale convolutional neural network for dynamic scene deblurring,'' in \emph{CVPR}, 2017.

\bibitem{wei2018deep}
C.~Wei, W.~Wang, W.~Yang, and J.~Liu, ``Deep retinex decomposition for low-light enhancement,'' in \emph{BMVC}, 2018.

\bibitem{TIAN2020461}
C.~Tian, Y.~Xu, and W.~Zuo, ``Image denoising using deep cnn with batch renormalization,'' \emph{Neural Networks}, vol. 121, pp. 461--473, 2020.

\bibitem{fu2019lightweight}
X.~Fu, B.~Liang, Y.~Huang, X.~Ding, and J.~Paisley, ``Lightweight pyramid networks for image deraining,'' \emph{IEEE transactions on neural networks and learning systems}, vol.~31, no.~6, pp. 1794--1807, 2019.

\bibitem{8750830}
Q.~Fan, D.~Chen, L.~Yuan, G.~Hua, N.~Yu, and B.~Chen, ``A general decoupled learning framework for parameterized image operators,'' \emph{IEEE Transactions on Pattern Analysis and Machine Intelligence}, vol.~43, no.~1, pp. 33--47, 2021.

\bibitem{8953950}
H.~Gao, X.~Tao, X.~Shen, and J.~Jia, ``Dynamic scene deblurring with parameter selective sharing and nested skip connections,'' in \emph{CVPR}, 2019.

\bibitem{Valanarasu2021TransWeatherTR}
J.~M.~J. Valanarasu, R.~Yasarla, and V.~M. Patel, ``Transweather: Transformer-based restoration of images degraded by adverse weather conditions,'' \emph{CVPR}, 2021.

\bibitem{liu2022tape}
L.~Liu, L.~Xie, X.~Zhang, S.~Yuan, X.~Chen, W.~Zhou, H.~Li, and Q.~Tian, ``Tape: Task-agnostic prior embedding for image restoration,'' in \emph{ECCV}, 2022.

\bibitem{zhang2023ingredient}
J.~Zhang, J.~Huang, M.~Yao, Z.~Yang, H.~Yu, M.~Zhou, and F.~Zhao, ``Ingredient-oriented multi-degradation learning for image restoration,'' in \emph{CVPR}, 2023.

\bibitem{wang2024gridformer}
T.~Wang, K.~Zhang, Z.~Shao, W.~Luo, B.~Stenger, T.~Lu, T.-K. Kim, W.~Liu, and H.~Li, ``Gridformer: Residual dense transformer with grid structure for image restoration in adverse weather conditions,'' \emph{International Journal of Computer Vision}, pp. 1--23, 2024.

\bibitem{Zamir2020MIRNet}
S.~W. Zamir, A.~Arora, S.~Khan, M.~Hayat, F.~S. Khan, M.-H. Yang, and L.~Shao, ``Learning enriched features for real image restoration and enhancement,'' in \emph{ECCV}, 2020.

\bibitem{Ren_2021_CVPR}
C.~Ren, X.~He, C.~Wang, and Z.~Zhao, ``Adaptive consistency prior based deep network for image denoising,'' in \emph{CVPR}, 2021.

\bibitem{mou2021gatir}
M.~Chong, Z.~Jian, and W.~Zhuoyuan, ``Dynamic attentive graph learning for image restoration,'' in \emph{ICCV}, 2021.

\bibitem{zafar2024single}
A.~Zafar, D.~Aftab, R.~Qureshi, X.~Fan, P.~Chen, J.~Wu, H.~Ali, S.~Nawaz, S.~Khan, and M.~Shah, ``Single stage adaptive multi-attention network for image restoration,'' \emph{IEEE Transactions on Image Processing}, 2024.

\bibitem{chu2022nafssr}
X.~Chu, L.~Chen, and W.~Yu, ``Nafssr: Stereo image super-resolution using nafnet,'' in \emph{CVPR Workshops}, 2022.

\bibitem{SIDD_2018_CVPR}
A.~Abdelhamed, S.~Lin, and M.~S. Brown, ``A high-quality denoising dataset for smartphone cameras,'' in \emph{CVPR}, 2018.

\end{thebibliography}

\end{document}